\pdfoutput=1
\documentclass[10pt,logo,copyright]{nvidiatechreport}
\usepackage[authoryear,sort&compress,round]{natbib}

\usepackage[utf8]{inputenc} % allow utf-8 input
\usepackage[T1]{fontenc}    % use 8-bit T1 fonts

\usepackage{parskip}        % no paragraph indents
\usepackage{url}            % simple URL typesetting
\usepackage{hyperref}       % hyperlinks
\usepackage{booktabs}       % professional-quality tables
\usepackage{amsfonts}       % blackboard math symbols
\usepackage{nicefrac}       % compact symbols for 1/2, etc.
\usepackage{microtype}      % microtypography
\usepackage{xcolor}         % colors
\usepackage[dvipsnames]{xcolor} % more color names
\usepackage{graphicx}
\usepackage{animate}        % for 360 video in teaser
\usepackage{subcaption}
\usepackage{tabularx}
\usepackage{makecell}
\usepackage{adjustbox}
\usepackage{setspace}
\newcolumntype{M}[1]{>{\centering\arraybackslash}m{#1}}
\usepackage{float}
\usepackage{tikz}
\usetikzlibrary{positioning,shapes,arrows}
\usepackage{amsmath,amsfonts,bm, bbm,leftindex}
\usepackage{multirow}
\usepackage{comment}
\usepackage{gensymb}
\usepackage{lipsum}
\usetikzlibrary{arrows.meta, positioning, fit}
\usepackage[para]{threeparttable}
\usepackage{tikz}
\usetikzlibrary{tikzmark}

\def\eqref#1{equation~\ref{#1}}
\def\1{\bm{1}}

\DeclareMathAlphabet{\mathsfit}{\encodingdefault}{\sfdefault}{m}{sl}
\SetMathAlphabet{\mathsfit}{bold}{\encodingdefault}{\sfdefault}{bx}{n}
\makeatletter
\let\save@mathaccent\mathaccent
\newcommand*\if@single[3]{%
  \setbox0\hbox{${\mathaccent"0362{#1}}^H$}%
  \setbox2\hbox{${\mathaccent"0362{\kern0pt#1}}^H$}%
  \ifdim\ht0=\ht2 #3\else #2\fi
  }
\newcommand*\rel@kern[1]{\kern#1\dimexpr\macc@kerna}
\newcommand*\widebar[1]{\@ifnextchar^{{\wide@bar{#1}{0}}}{\wide@bar{#1}{1}}}
\newcommand*\wide@bar[2]{\if@single{#1}{\wide@bar@{#1}{#2}{1}}{\wide@bar@{#1}{#2}{2}}}
\newcommand*\wide@bar@[3]{%
  \begingroup
  \def\mathaccent##1##2{%
    \let\mathaccent\save@mathaccent
    \if#32 \let\macc@nucleus\first@char \fi
    \setbox\z@\hbox{$\macc@style{\macc@nucleus}_{}$}%
    \setbox\tw@\hbox{$\macc@style{\macc@nucleus}{}_{}$}%
    \dimen@\wd\tw@
    \advance\dimen@-\wd\z@
    \divide\dimen@ 3
    \@tempdima\wd\tw@
    \advance\@tempdima-\scriptspace
    \divide\@tempdima 10
    \advance\dimen@-\@tempdima
    \ifdim\dimen@>\z@ \dimen@0pt\fi
    \rel@kern{0.6}\kern-\dimen@
    \if#31
      \overline{\rel@kern{-0.6}\kern\dimen@\macc@nucleus\rel@kern{0.4}\kern\dimen@}%
      \advance\dimen@0.4\dimexpr\macc@kerna
      \let\final@kern#2%
      \ifdim\dimen@<\z@ \let\final@kern1\fi
      \if\final@kern1 \kern-\dimen@\fi
    \else
      \overline{\rel@kern{-0.6}\kern\dimen@#1}%
    \fi
  }%
  \macc@depth\@ne
  \let\math@bgroup\@empty \let\math@egroup\macc@set@skewchar
  \mathsurround\z@ \frozen@everymath{\mathgroup\macc@group\relax}%
  \macc@set@skewchar\relax
  \let\mathaccentV\macc@nested@a
  \if#31
    \macc@nested@a\relax111{#1}%
  \else
    \def\gobble@till@marker##1\endmarker{}%
    \futurelet\first@char\gobble@till@marker#1\endmarker
    \ifcat\noexpand\first@char A\else
      \def\first@char{}%
    \fi
    \macc@nested@a\relax111{\first@char}%
  \fi
  \endgroup
}
\makeatother
\newcommand{\modelname}{GR00T N1}

\newcommand{\modellarge}{GR00T-N1-2B}

\newcommand{\FIGREF}[1]{Fig.~\ref{#1}} % Fig. or Figure
\definecolor{darkred}{rgb}{0.7, 0.0, 0.0}

\usepackage{pifont}
\usepackage{wrapfig}

\usepackage[nameinlink]{cleveref}
\crefname{equation}{Eq.}{Eqs.}
\crefname{figure}{Fig.}{Figs.}
\crefname{section}{Sec.}{Sec.}
\crefname{appendix}{App.}{App.}
\crefname{table}{Tab.}{Tabs.}
\crefname{algorithm}{Algo}{Algo}
\crefname{thm}{Thm}{Thm}
\Crefname{thm}{Thm}{Thm}
\crefname{prop}{Prop}{Prop}
\usepackage{longtable}

\newcommand{\crefnames}[3]{%
  \@for\next:=#1\do{%
    \expandafter\crefname\expandafter{\next}{#2}{#3}%
  }%
}

\title{\modelname{}: An Open Foundation Model for Generalist Humanoid Robots}

\author{NVIDIA\footnote{A detailed list of contributors and acknowledgments can be found in~\cref{sec:contributors} of this paper.}}

\begin{abstract}
General-purpose robots need a versatile body and an intelligent mind. Recent advancements in humanoid robots have shown great promise as a hardware platform for building generalist autonomy in the human world. A robot foundation model, trained on massive and diverse data sources, is essential for enabling the robots to reason about novel situations, robustly handle real-world variability, and rapidly learn new tasks. To this end, we introduce \modelname{}, an open foundation model for humanoid robots. \modelname{} is a Vision-Language-Action (VLA) model with a dual-system architecture. The vision-language module (System 2) interprets the environment through vision and language instructions. The subsequent diffusion transformer module (System 1) generates fluid motor actions in real time. Both modules are tightly coupled and jointly trained end-to-end. We train \modelname{} with a heterogeneous mixture of real-robot trajectories, human videos, and synthetically generated datasets.
We show that our generalist robot model \modelname{} outperforms the state-of-the-art imitation learning baselines on standard simulation benchmarks across multiple robot embodiments. Furthermore, we deploy our model on the Fourier GR-1 humanoid robot for language-conditioned bimanual manipulation tasks, achieving strong performance with high data efficiency.
\end{abstract}

\begin{document}

\maketitle

\abscontent

% \checkthis{Timeline: Initial draft due 5pm PT Tuesday (2/25).}
% \checkthis{If your names are in \textbf{bold}, it means that your assigned sections are required for the initial draft. }

\section{Introduction}
\label{sec:intro}

Creating autonomous robots to perform everyday tasks in the human world has long been a fascinating goal and, at the same time, a significant technical undertaking. Recent progress in robotic hardware, artificial intelligence, and accelerated computing has collectively paved the ground for developing general-purpose robot autonomy. To march toward human-level physical intelligence, we advocate for a full-stack solution that integrates the three key ingredients: hardware, models, and data. First and foremost, robots are embodied physical agents, and their hardware determines their capability envelope. It makes \textbf{humanoid robots} a compelling form factor to build robot intelligence due to their human-like physique and versatility. Second, the diversity and variability of the real world demands that the robots operate on open-ended objectives and perform a wide range of tasks. Achieving this requires a \textbf{generalist robot model} sufficiently expressive and capable of handling various tasks. Third, real-world humanoid data are costly and time-consuming to acquire at scale. We need an effective data strategy to train large-scale robotic models.

In recent years, foundation models~\cite{} have brought forth dramatic breakthroughs in understanding and generating visual and text data. They demonstrate the effectiveness of training generalist models on web-scale data to enable strong generalization and fast adaptation to downstream tasks. The successes of foundation models in neighboring fields of AI have depicted a promising roadmap for building the ``backbone'' of intelligence for generalist robots, endowing them with a set of core competencies and enabling them to rapidly learn and adapt in the real world.
%so where is the foundation model of \emph{actions} for humanoid robot control?
%
However, unlike the digital realms of words and pixels, no Internet of humanoid robot datasets exist for large-scale pre-training. The data available for any single humanoid hardware would be orders of magnitude too small. Recent efforts in the robot learning community~\citep{open_x_embodiment_rt_x_2023} have explored cross-embodied learning to enlarge the dataset by pooling training data from many different robots. However, the great variability in robot embodiments, sensors, actuator degrees of freedom, control modes, and other factors result in an archipelago of ``data islands'' rather than a coherent, Internet-scale dataset needed for training a true generalist model.

\begin{wrapfigure}{R}{0.5\textwidth}
\begin{center}
\vspace{-4mm}
\includegraphics[height=5cm]{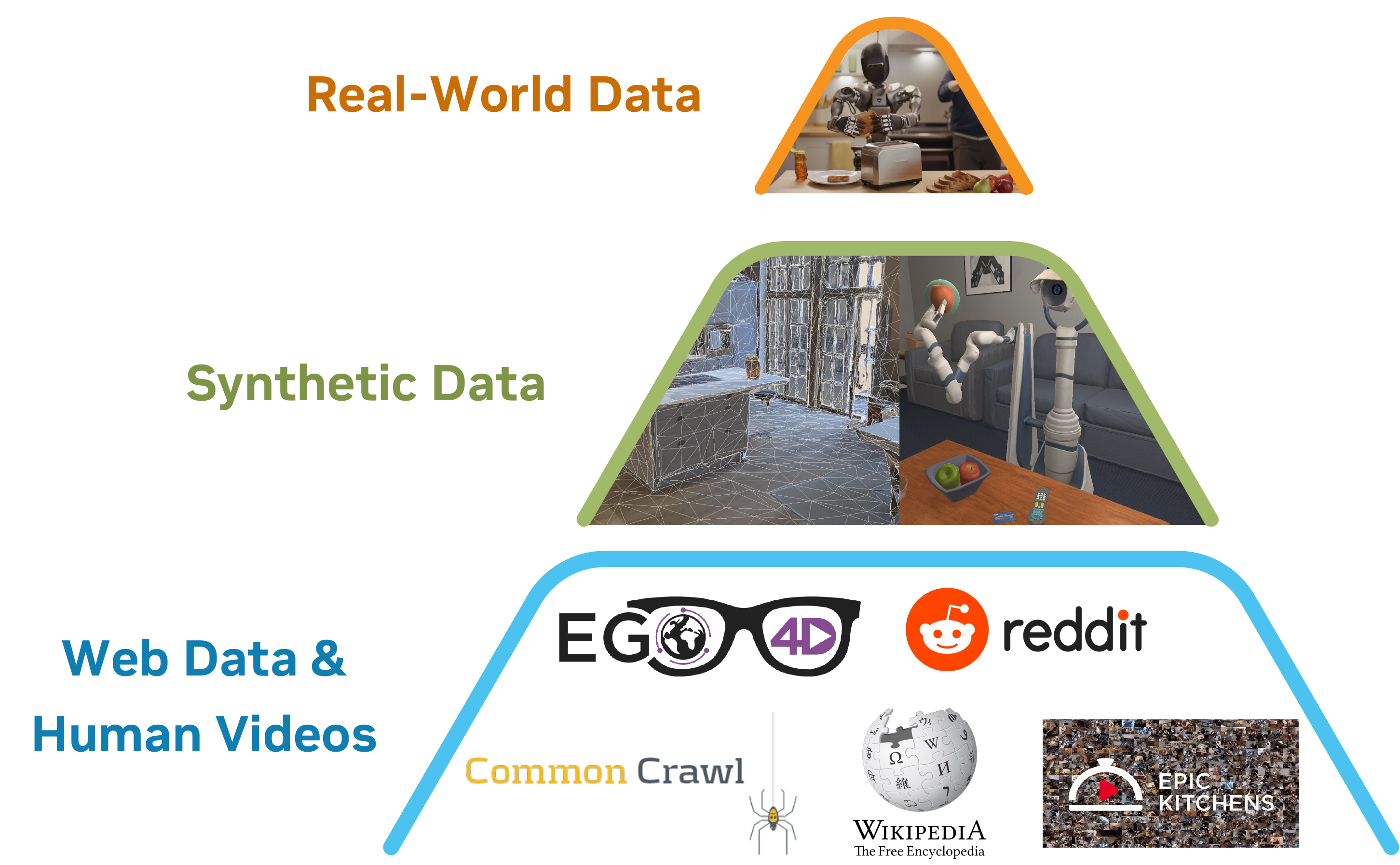}
\caption{\textbf{Data Pyramid for Robot Foundation Model Training.} \modelname{}'s heterogeneous training corpora can be represented as a pyramid: data quantity decreases, and embodiment-specificity increases, moving from the bottom to the top.\label{fig:data_pyramid}} 
%\checkthis{can we rename parts in the image/section titles in \Cref{sec:datasets} so that they match?}}
\end{center}
\end{wrapfigure}

We introduce \modelname{}, an open foundation model for generalist humanoid robots. The \modelname{} model is a Vision-Language-Action (VLA) model, which generates actions from image and language instruction input. It has cross-embodiment support from tabletop robot arms to dexterous humanoid robots. It adopts a \textbf{dual-system compositional architecture}, inspired by human cognitive processing~\citep{kahneman2011thinking}. The System 2 reasoning module is a pre-trained Vision-Language Model (VLM) that runs at 10Hz on an NVIDIA L40 GPU. It processes the robot's visual perception and language instruction to interpret the environment and understand the task goal. Subsequently, a Diffusion Transformer, trained with action flow-matching, serves as the System 1 action module. It cross-attends to the VLM output tokens and employs embodiment-specific encoders and decoders to handle variable state and action dimensions for motion generation. It generates closed-loop motor actions at a higher frequency (120Hz). Both the System 1 and System 2 modules are implemented as Transformer-based neural networks, tightly coupled and jointly optimized during training to facilitate coordination between reasoning and actuation.

To mitigate the ``data island'' problem mentioned earlier, we structure the VLA training corpora as a \textbf{data pyramid}, illustrated in \FIGREF{fig:data_pyramid}. Rather than treating the training datasets as a homogeneous pool, we organize heterogeneous sources by scale: large quantities of web data and human videos lay the base of the pyramid; synthetic data generated with physics simulations and/or augmented by off-the-shelf neural models form the middle layer, and real-world data collected on the physical robot hardware complete the top. The lower layers of the pyramid provide broad visual and behavioral priors, while the upper layers ensure grounding in embodied, real-robot execution.

% To harness the entire data pyramid for model training, 
We develop an effective \textbf{co-training} strategy to learn across the entire data pyramid in both pre- and post-training phases. To train our model with action-less data sources, such as human videos and neural-generated videos, we learn a latent-action codebook~\citep{ye2025latent} and also use a trained inverse dynamics model (IDM) to infer pseudo-actions. These techniques enable us to annotate actions on action-less videos so we can effectively treat them as additional robot embodiments for model training. By unifying all data sources across the data pyramid, we construct a consistent dataset where the input consists of the robot state, visual observations, and language instruction, and the output is the corresponding motor action. We pre-train our model end-to-end across the three data layers, spanning (annotated) video datasets, synthetically generated datasets, and real-robot trajectories --- by sampling training batches across this heterogeneous data mixture.

With a unified model and single set of weights, \modelname{} can generate diverse manipulation behaviors using single-arm, bimanual, and humanoid embodiments.
Evaluated on standard simulation benchmark environments, \modelname{} achieves superior results compared to state-of-the-art imitation learning baselines. We also demonstrate \modelname{}'s strong performance in real-world experiments with GR-1 humanoid robots. Our \modellarge{} model checkpoint, training data, and simulation benchmarks are publicly available here: \href{http://github.com/NVIDIA/Isaac-GR00T}{GitHub} and \href{https://huggingface.co/datasets/nvidia/PhysicalAI-Robotics-GR00T-X-Embodiment-Sim}{HuggingFace Datasets}.

\section{\modelname{} Foundation Model}
\label{sec:model}

% \textbf{@Scott Reed (PIC), @Jim Fan, @Yuke Zhu}

\begin{figure}[t]
\includegraphics[width=\textwidth]{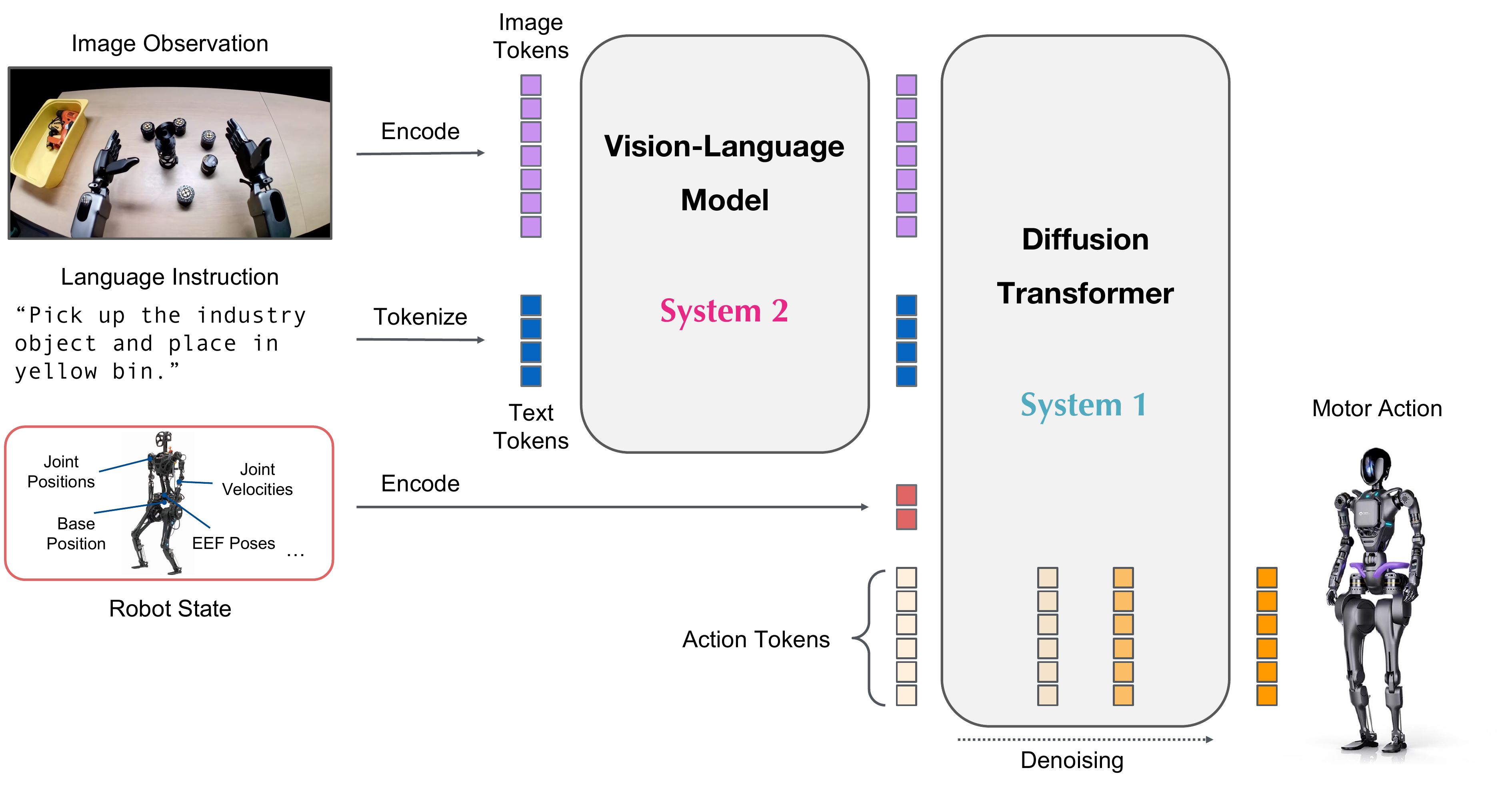}
\caption{\textbf{\modelname{} Model Overview.} Our model is a Vision-Language-Action (VLA) model that adopts a dual-system design.  We convert the image observation and language instruction into a sequence of tokens to be processed by the Vision-Language Model (VLM) backbone. The VLM outputs, together with robot state and action encodings, are passed to the Diffusion Transformer module to generate motor actions.
\label{fig:groot_inference}}
\end{figure}

\modelname{} is a Vision-Language-Action (VLA) model for humanoid robots trained on diverse data sources. 
The model contains a vision-language backbone that encodes language and image input and a DiT-based flow-matching policy that outputs high-frequency actions.
We use the NVIDIA Eagle-2 VLM~\citep{eagle2} as the vision-language backbone. 
Specifically, our publicly released \modellarge{} model has 2.2B parameters in total, with 1.34B in the VLM.
The inference time for sampling a chunk of 16 actions is 63.9ms on an L40 GPU using bf16.
\FIGREF{fig:groot_inference} provides a high-level overview of our model design. We highlight three key features of \modelname{}:

\begin{itemize}
\item We design a compositional model that integrates Vision-Language Model (VLM)-based reasoning module (System 2) and Diffusion Transformer (DiT)-based action module (System 1) in a unified learning framework;
\item We develop an effective pre-training strategy using a mixture of human videos, simulation and neural-generated data, and real robot demonstrations (see \FIGREF{fig:data_pyramid}) for generalization and robustness;
\item We train a massively multi-task, language-conditioned policy that supports a wide range of robot embodiments and enables rapid adaptation to new tasks through data-efficient post-training.
\end{itemize}

\subsection{Model Architecture}
\label{sec:model_architecture}
\begin{figure}[t]
\includegraphics[width=1.0\linewidth]{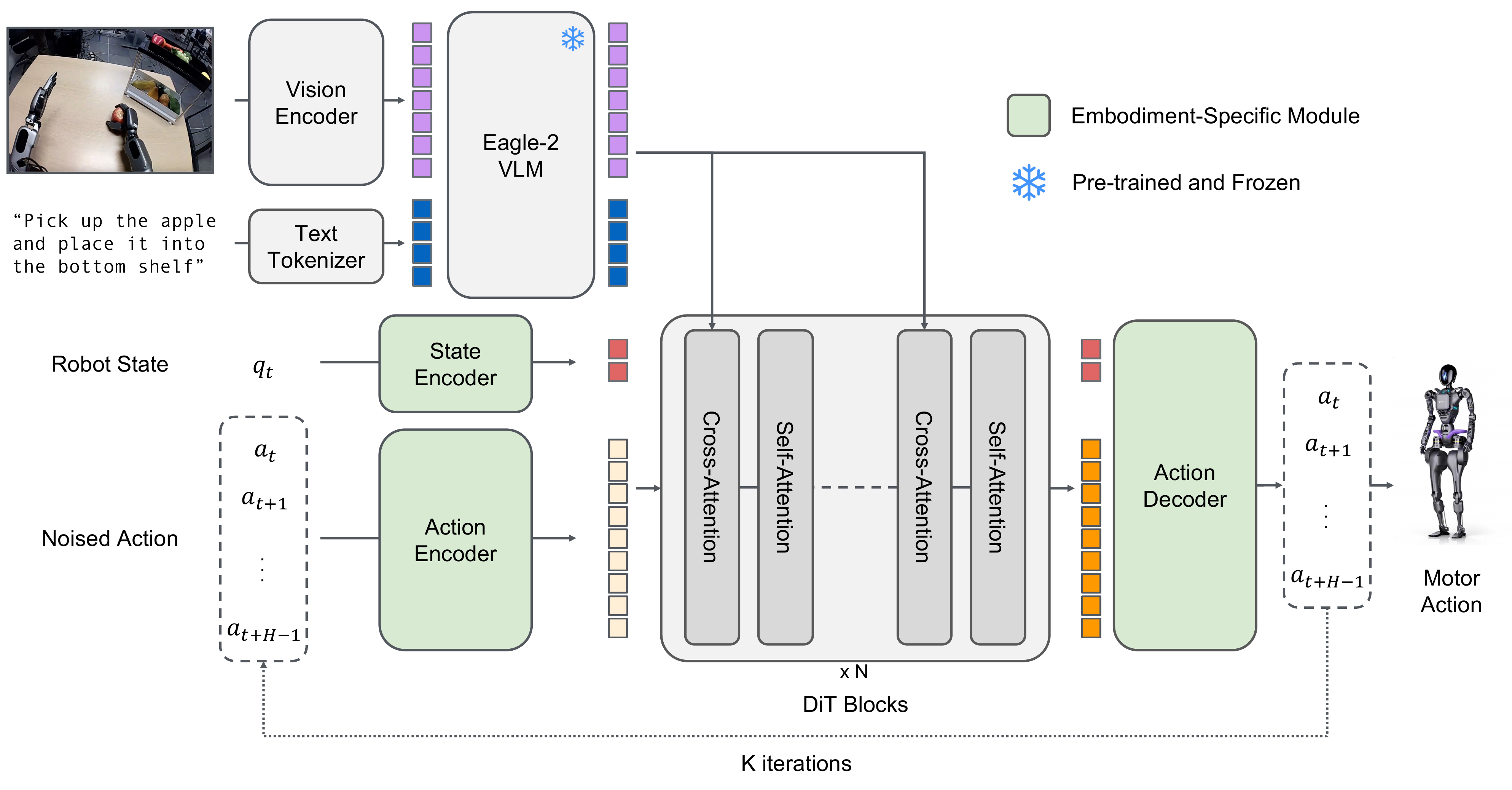}
\caption{\textbf{\modelname{} Model Architecture.} \modelname{} is trained on a diverse set of embodiments ranging from single-arm robot arms to bimanual humanoid dexterous hands. 
To deal with different robot embodiment's state observation and action, we use DiT blocks with an embodiment-aware state and action encoder to embed the robot's state and action inputs.
\modelname{} model leverages latent embeddings of the Eagle-2 model to incorporate the robot's visual observation and language instructions. The vision language tokens will then be fed into the DiT blocks through cross-attention layers. \label{fig:network_architecture}}
\end{figure}

In this section, we describe the \modelname{} model architecture, illustrated in \FIGREF{fig:network_architecture}.
\modelname{} uses flow-matching \citep{flowmatching} to learn action generation.
A diffusion transformer (DiT) processes the robot's proprioceptive state and action, which are then cross-attended with image and text tokens from the Eagle-2 VLM backbone to output the denoised motor actions. Below, we elaborate on each module in detail.

\paragraph{State and Action Encoders} 
To process states and actions of varying dimensions across different robot embodiments, we use an MLP per embodiment to project them to a shared embedding dimension as input to the DiT.
As in~\citet{black2024pi_0}, the Action Encoder MLP also encodes the diffusion timestep together with the noised action vector.

We use action flow matching, which samples actions through iterative denoising. The model takes as input noised actions in addition to encodings of the robot's proprioceptive state, image tokens, and text tokens.
The actions are processed in chunks as in~\citet{action_chunking}, meaning that at any given time $t$ the model uses $A_t = [a_t, a_{t+1}, \ldots, a_{t+H-1}]$ which contains the action vectors of timesteps $t$ through $t+H -1$. 
We set $H=16$ in our implementation.

\paragraph{Vision-Language Module (System 2)}
For encoding vision and language inputs, \modelname{} uses the Eagle-2~\citep{eagle2} vision-language model (VLM) pretrained on Internet-scale data. 
Eagle-2 is finetuned from a SmolLM2 \citep{allal2025smollm2} LLM and a SigLIP-2 \citep{tschannen2025siglip} image encoder. 
Images are encoded at resolution $224 \times 224$ followed by pixel shuffle \citep{pixel_shuffle}, resulting in 64 image token embeddings per frame. These embeddings are then further encoded together with text by the LLM component of the Eagle-2 VLM.
The LLM and image encoder are aligned over a broad set of vision-language tasks following the general recipe of \citet{eagle2}. 

During policy training, a text description of the task, as well as (possibly multiple) images, are passed to the VLM in the chat format used during vision-language training. We then extract vision-language features of shape (batch size $\times$ sequence length $\times$ hidden dimension) from the LLM. 
We found that using middle-layer instead of final-layer LLM embeddings resulted in both faster inference speed and higher downstream policy success rate.
For \modellarge{}, we use the representations from the 12th layer. 

\paragraph{Diffusion Transformer Module (System 1)}
For modeling actions, \modelname{} uses a variant of DiT~\citep{peebles2023scalable}, which is a transformer with denoising step conditioning via adaptive layer normalization, denoted as $V_\theta$.
As shown in \FIGREF{fig:network_architecture}, $V_\theta$ consists of alternating cross-attention and self-attention blocks, similar to Flamingo~\citep{NEURIPS2022_960a172b} and VIMA~\citep{jiang2023vima}.
The self-attention blocks operate on noised action token embeddings $A_t^{\tau}$ together with state embeddings $q_t$, while cross-attention blocks allow conditioning on the vision-language token embeddings $\phi_t$ output by VLM.
After the final DiT block, we apply an embodiment-specific Action Decoder, another MLP, to the final $H$ tokens to predict the actions.

Given a ground-truth action chunk $A_t$, a flow-matching timestep $\tau \in [0, 1]$ and sampled noise $\epsilon \sim \mathcal{N}(\mathbf{0}, \mathbf{I})$, the noised action chunk $A_t^{\tau}$ is computed as $A_t^{\tau} = \tau A_t + (1-\tau) \epsilon$. 
The model prediction $V_\theta(\phi_t, A_t^{\tau}, q_t)$ aims to approximate the denoising vector field $\epsilon - A_t$ by minimizing the following loss:
\begin{equation}
\label{eq:fm_loss}
\mathcal{L}_{\textit{fm}}(\theta) = \mathbb{E}_{\tau} \left[\| V_\theta(\phi_t, A_t^{\tau}, q_t) - (\epsilon - A_t)\|^2\right].
\end{equation}
As in~\citet{black2024pi_0}, we use $p(\tau) = \text{Beta}(\frac{s-\tau}{s}; 1.5, 1)$, $s=0.999$.
During inference, we generate action chunks with $K$-step denoising.
First, randomly sample $A_t^0 \sim \mathcal{N}(\mathbf{0}, \mathbf{I})$ and then use forward Euler integration to iteratively generate the action chunk, updating as follows:
$$
A_t^{\tau + 1/K} = A_t^{\tau} + \frac{1}{K} V_\theta(\phi_t, A_t^{\tau}, q_t).
$$
In practice, we found $K=4$ inference steps to work well across all embodiments.

\subsection{Training Data Generation}
\label{sec:training_data_generation}
\begin{figure}[thb!]
\centering
\includegraphics[width=0.95\textwidth]{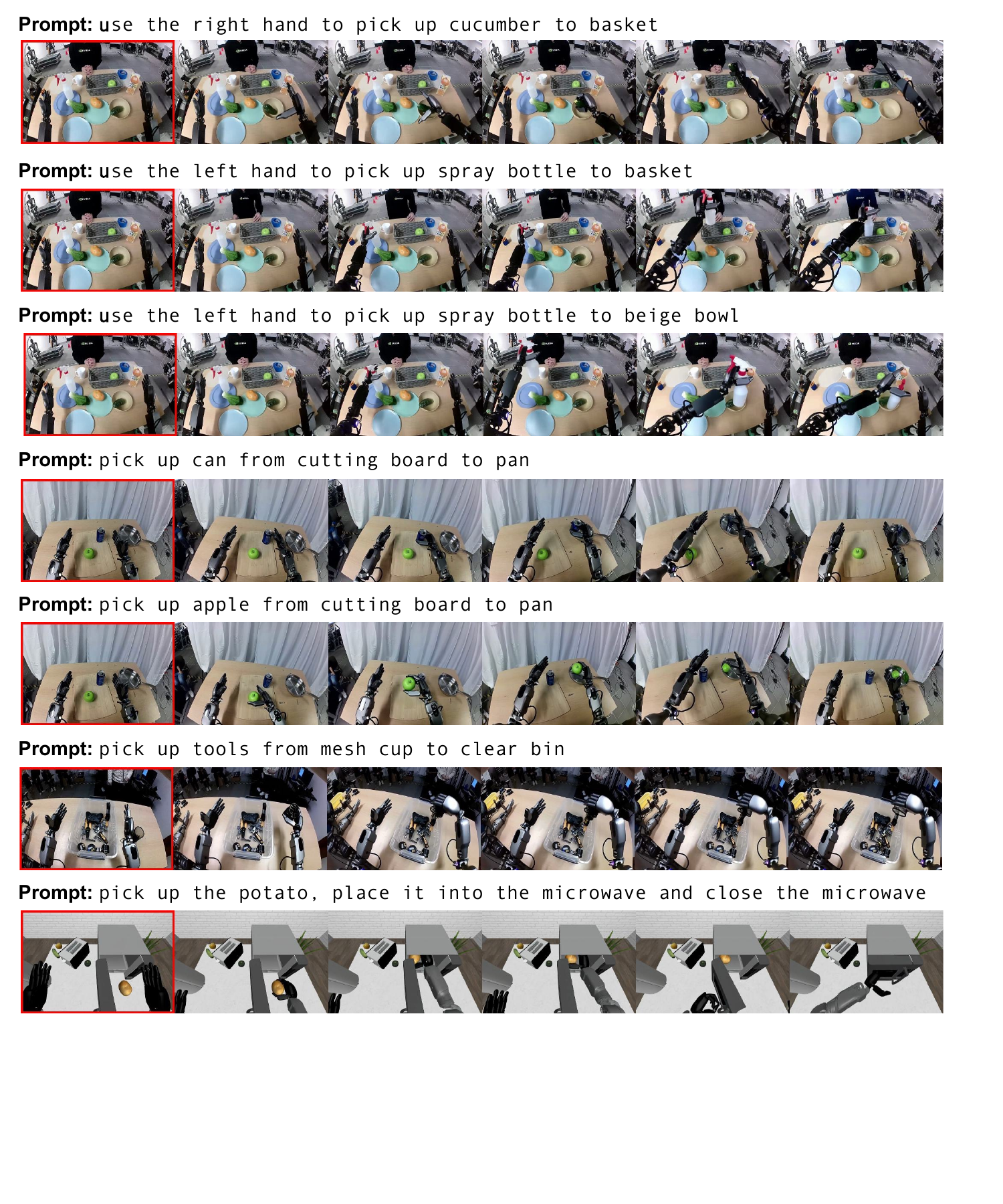}
\vspace{-8em}
\caption{\textbf{Synthetically Generated Videos.} We leverage off-the-shelf video generation models to create neural trajectories to increase the quantity and diversity of our training datasets. These generated data can be used for both pre- and post-training of our \modelname{}. (1) The first three rows are generated from the same initial frames but with different prompts (change left or right, the location to place the object), (2) the following two are from the same initial frames but replace the object to pick up, (3) the next row showcases the video model generating a robot trajectory which is very challenging to generate in simulation (spilling contents inside a mesh cup into a bin), and (4) the last row is generated from an initial frame from simulation data. We use the red rectangles to indicate the initial frames.}
\label{fig:dream_sample}
\end{figure}

To train \modelname{}, we use a diverse set of data sources and objectives to construct the data pyramid (\FIGREF{fig:data_pyramid}). 
We first source diverse human egocentric video data from open datasets, which forms the base, together with the web data used in VLM pretraining.
Next, we generate synthetic \textit{neural} trajectories using pre-trained video generation models. 
In this way, we $\sim$10$\times$ our in-house collected teleoperation trajectories --- the ``peak'' of the data pyramid --- from 88 hours to 827 hours, using diverse counterfactual robot trajectories with novel language instructions (see \FIGREF{fig:dream_sample} for examples). We additionally generate diverse simulation trajectories, which also expand the middle part of the data pyramid.

In the next paragraph, we first describe how we extract \textit{latent} actions from videos, which we use to extract labels for web-scaled human egocentric datasets. Next, we describe how we generate \textit{neural} and \textit{simulated} robot trajectories, and how we obtain actions for each of these data sources.

\begin{figure}[thb!]
\centering
\includegraphics[width=0.95\textwidth]{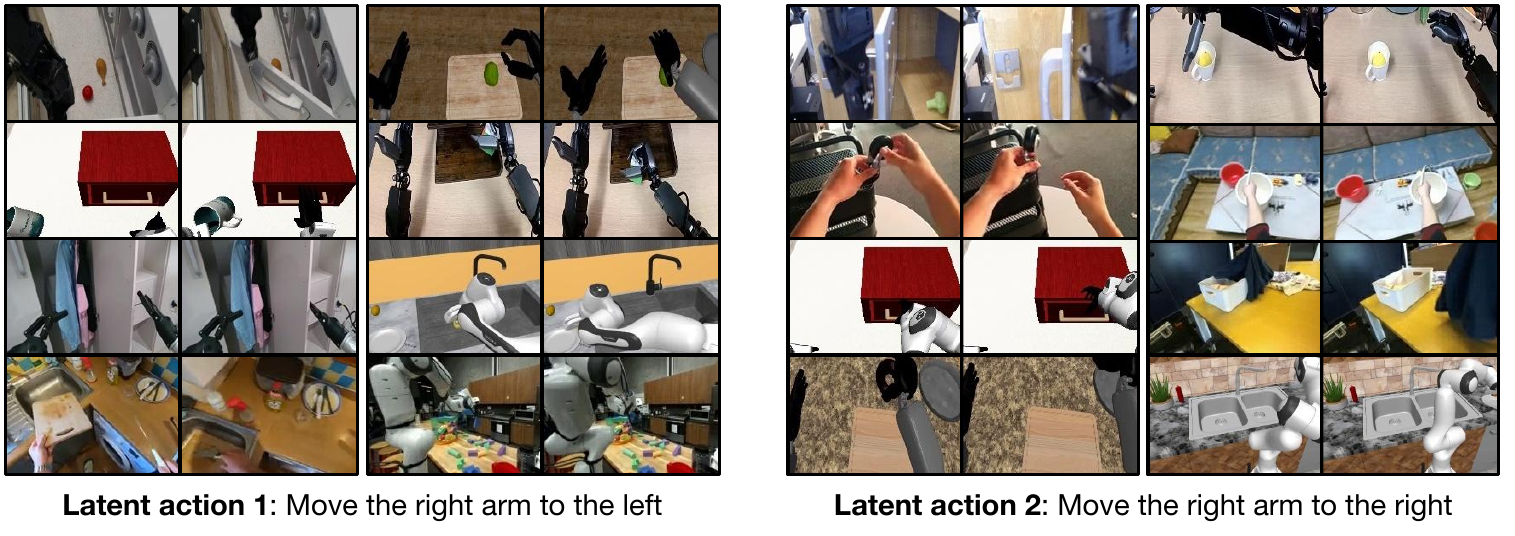}
% \vspace{-8em}
\caption{\textbf{Latent Actions.} We retrieve similar latent embeddings across various embodiments. The left images illustrate the latent action that corresponds to moving the right arm (or hand) to the left, while the right images illustrate the latent action that corresponds to moving the right arm (or hand) to the right. Note that this general latent action is not only consistent in different robot embodiments, but also in human embodiment.}
\label{fig:latent_sample}
\end{figure}

\paragraph{Latent Actions} 
For human egocentric videos and neural trajectories, we do not have any actions that we can directly use to train \modelname{}. 
For these data, we instead generate latent actions by training a VQ-VAE model to extract features from consecutive image frames from videos~\citep{ye2025latent}.
The encoder takes the current frame $x_t$ and the future frame $x_{t+H}$ of a video with a fixed window size $H$ and outputs the latent action $z_t$.
The decoder is trained to take the latent action $z_t$ and $x_t$ and reconstruct $x_{t+H}$. This model is trained with a VQ-VAE objective, where the continuous embedding from the encoder is mapped to the nearest embedding from the codebook. 
After training, we take the encoder and use it as an inverse dynamics model; given an $x_t$ and $x_{t+H}$ pair, we extract the continuous pre-quantized embedding and use this as the latent action label during pre-training, with the same flow-matching loss, but treat it as a distinct ``LAPA'' embodiment.
Training the VQ-VAE model on all heterogeneous data together allows us to unify all of the data to share the same learned latent action space, potentially improving cross-embodiment generalization. \FIGREF{fig:latent_sample} shows $x_t$ and $x_{t+H}$ pairs from 8 distinct embodiments including both robot and human embodiment, all retrieved from similar latent actions; the first latent action shows all embodiments \textit{moving right arm to the left} and the second latent action shows \textit{moving right arm to the right}.

\paragraph{Neural Trajectories} 
\label{sec:neural_trajectories}
Robot data scales linearly with human labor, since it typically requires a human operator to teleoperate the robot to produce each trajectory. 
Recently, \textbf{video generation models} have demonstrated significant potential for high-quality controllable video generation~\citep{brooks2024video, yang2024cogvideox, xiang2024pandora, lin2024stiv, wan2.1, ren2025videoworld}, which paves the way for building world models in the robotic domain. 
To harness these models, we fine-tune image-to-video generation models~\citep{agarwal2025cosmos, yang2024cogvideox, wan2.1} on all of our 88 hours of in-house collected teleoperation data and generate 827 hours of video data given the existing initial frames with novel language prompts, augmenting it by around 10$\times$. 
This enables generating training data that captures many more counterfactual scenarios in the real world without actually collecting teleoperation data for each of these cases (examples shown in \FIGREF{fig:dream_sample}; more examples of dream generations in \FIGREF{fig:appendix-dream}). 

To increase the diversity of our neural trajectories, we first use a commercial-grade multimodal LLM to detect the objects given initial frames and generate many more possible combinations of ``\textit{pick up \{object\} from \{location A\} to \{location B\}}", while instructing the model to only consider the physically feasible combinations.
We also apply post-processing mechanisms, including filtering and re-captioning, to the generated videos. 
For this, we also use a commercial-grade multimodal LLM as a judge and feed the downsampled 8 frames to filter out neural trajectories that do not follow the language instruction precisely.
% that 1) do not follow the language instruction or 2) have inaccurate physics \checkthis{(Appendix \ref{} shows some examples of these failure cases)}. 
%
We then caption the filtered-out videos. (More details can be found in Appendix~\ref{sec:appendix_dream}).

\paragraph{Simulation Trajectories}
Scaling up real-world data collection for humanoid robots is highly expensive due to the challenge of simultaneously controlling both arms and dexterous hands.
Recent research~\citep{wang2023robogen,mandlekar2023mimicgen,jiang2024dexmimicen} has demonstrated that generating training data in simulation is a practical alternative.
We use DexMimicGen~\citep{jiang2024dexmimicen} to synthesize large-scale robot manipulation trajectories.

Starting with a small set of human demonstrations, DexMimicGen applies demonstration transformation and replay in simulation to expand the dataset automatically. 
Each task is decomposed into a sequence of object-centric subtasks. 
The initial human demonstrations are segmented into smaller manipulation sequences, each corresponding to a subtask involving a single object. 
These segments are then adapted to new environments by aligning them with the object's position, preserving the relative poses between the robot's end effector and the object.
To ensure smooth execution, the system interpolates movements between the robot's current state and the transformed segment.
The robot then follows the full sequence step by step, verifying task success at the end. Only successful demonstrations are retained, ensuring high-quality data. Using DexMimicGen, we scale a limited set of human demonstrations into a large-scale humanoid manipulation dataset. Considering the pre- and post-training datasets, we have generated 780,000 simulation trajectories --- equivalent to 6,500 hours, or nine continuous months, of human demonstration data --- in just 11 hours. These simulation data significantly supplement the real-robot data with minimal human costs.

\subsection{Training Details}
\label{sec:training_details}

\paragraph{Pre-training}

During the pre-training phase, \modelname{} is trained via flow-matching loss (Equation~\ref{eq:fm_loss}) on a diverse collection of embodiments and data sources, encompassing various real and synthetic robot datasets as well as human motion data.
We refer readers to Sec.~\ref{sec:datasets} for a detailed description of the datasets.

For human videos, in the absence of ground-truth actions, we extract learned latent actions and use them as flow-matching targets (see Sec.~\ref{sec:training_data_generation}).
For robot datasets such as our GR-1 humanoid data or Open X-Embodiment data, we use both ground-truth robot actions as well as learned latent actions as flow-matching targets.
In the case of neural trajectories (Sec.~\ref{sec:neural_trajectories}) used to augment our robot datasets, we use both latent actions as well as predicted actions from an inverse-dynamics model trained on the real robot data.
Pre-training hyper-parameters are listed in Table~\ref{tab:hyperparameters} in the Appendix.

\paragraph{Post-training}

In the post-training phase, we fine-tune our pre-trained model on datasets corresponding to each single embodiment.
As in pretraining, we keep the language component of the VL backbone frozen and fine-tune the rest of the model.
Post-training hyperparameters are given in Table~\ref{tab:hyperparameters} in the Appendix.

\paragraph{Post-training with Neural Trajectories}
To overcome the challenge of data scarcity during post-training, we explore augmenting the data for each downstream task by generating neural trajectories, similar to the procedure described in Sec.~\ref{sec:training_data_generation}.
For downstream tasks that are conditioned on multiple views, we finetune the video model to generate multiple subimages in a grid (\FIGREF{fig:appendix-dream}).
For simulation tasks, we collect diverse initial frames from the randomly initialized environment. 
For real robot tasks, we randomly initialize object poses manually and record the robot's initial observation. 
Novel initial frames could also be created automatically using img2img diffusion (example shown in \FIGREF{fig:appendix-dream}), but we leave further exploration for future work.
We also demonstrate examples of (1) multi-round video generation for generating long-horizon trajectories composed of atomic tasks and (2) neural trajectories of liquids and articulated objects, known to be extremely challenging to simulate, though we leave quantitative evaluation of downstream tasks for future work.

For our post-training pipeline with neural trajectories, we restrict ourselves to fine-tuning the video generation model \textit{only} on the human-collected trajectories for simulation tasks and only 10\% of the data from the real-world benchmark collected for post-training, to match the realistic scenario that we only have access to limited number of teleoperation data. 
%While we usually only 10x our post-training data for this paper, we can infinitely generate neural trajectories given initial frames of the environment.
%
Since the generated videos do not have action labels, we use either latent or inverse dynamics models (IDM) labeled actions~\citep{baker2022video} and train the policy model to treat these pseudo-actions as action labels for a different embodiment. In low-data regime scenarios, we also restrict ourselves on training the IDM models only on the low-data, to facilitate realistic scenarios. Details of how we train the IDM models are provided in Appendix \ref{sec:appendix_idm}. Some empirical comparisons between latent and IDM-labeled actions are made in Sec. \ref{sec:quant_results}. 
During post-training, we co-train the policy with real-world trajectories with neural trajectories with a 1:1 sampling ratio. 

\paragraph{Training Infrastructure}

We train \modelname{} on a cluster managed via NVIDIA OSMO \citep{nvidia_osmo}, an orchestration platform for scaling complex ‌robotics workloads. The training cluster is equipped with H100 NVIDIA GPUs connected via NVIDIA Quantum-2 InfiniBand in a fat-tree topology. We facilitate fault-tolerant multi-node training and data ingestion via a custom library built on top of the Ray distributed computing library \citep{moritz2018ray}. We use up to 1024 GPUs for a single model. \modellarge{} used roughly 50,000 H100 GPU hours for pretraining.

Compute-constrained finetuning was tested in the context of a single A6000 GPU.
If only tuning the adapter layers (action and state encoders + action decoder) and DiT, a batch size up to 200 can be used.
When tuning the vision encoder, a batch size of up to 16 can be used.

\section{Pre-Training Datasets}
\label{sec:datasets}

We structure our pre-training corpus into three main categories: real-robot datasets (Sec.~\ref{datasets:real-world-datasets}), synthetic datasets (Sec.~\ref{datasets:synthetic-datasets}), and human video datasets (Sec.~\ref{datasets:human-video-datasets}).
These roughly correspond to the peak, middle, and base of the data pyramid (\FIGREF{fig:data_pyramid}), respectively. The synthetic datasets consist of both simulation trajectories and neural trajectories. Table~\ref{tab:synthesized_datasets} summarizes our training data generation strategies in Sec.~\ref{sec:training_data_generation} and their applicable data sources correspondingly. We provide the full statistics (\# of frames, hours, and camera views) of our pretraining datasets in Table \ref{tab:dataset_stats}.
%
%Each category plays a distinct role in providing the diverse, high-quality data necessary to build a generalist agent.

%This structure naturally aligns with the data pyramid concept (Fig.~\ref{fig:data_pyramid}): latent pre-training datasets form the broad base, consisting of the largest volume of data without action labels; action-based pre-training datasets occupy the middle layer, incorporating structured robotic demonstrations and human data with pseudo-actions; and post-training datasets sit at the peak, comprising high-quality robotic data in both simulation and real-world settings. This organization ensures a strategic balance between data quantity, generalization, and task specificity.

\begin{table}[htb]
    \centering
    \caption{\textbf{Training Data Generation.} Our data generation strategies leverage different data sources. The latent-action learning technique is broadly applied to diverse video datasets. Neural trajectories can be generated from datasets containing robot actions, while simulation trajectories rely on a physics simulator and utilize our DexMimicGen-based automated data generation system.}
    \label{tab:synthesized_datasets}
    \begin{tabular}{lccc}
    \toprule
         & Latent Actions & Neural Trajectories & Simulation Trajectories \\ \midrule
       Real-Robot Datasets & \checkmark & \checkmark
 & \checkmark
\\
       Simulated Robot Datasets & \checkmark
 & \checkmark
 & 
\\
       Human Video Datasets & \checkmark  & & 
\\
    \bottomrule
    \end{tabular}

\end{table}

\subsection{Real-World Datasets}
\label{datasets:real-world-datasets}

\begin{figure}[t]
\centering
\includegraphics[width=1.0\textwidth]{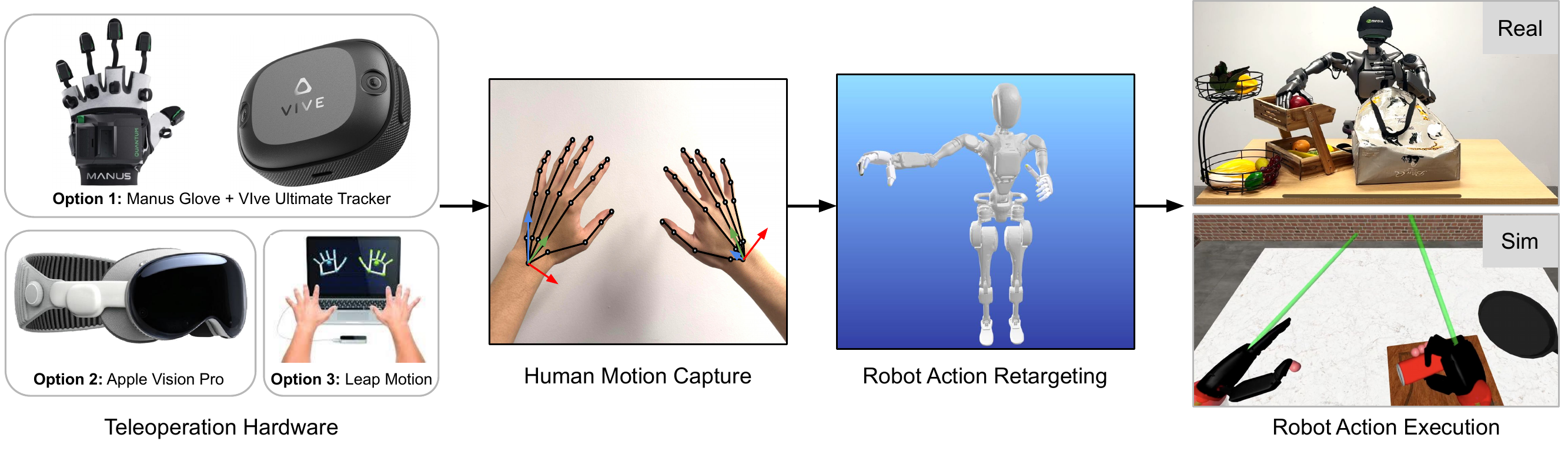}
\caption{\textbf{Data Collection via Teleoperation.} Our teleoperation infrastructure supports multiple devices to capture human hand motion, including 6-DoF wrist poses and hand skeletons. Robot actions are produced through retargeting and executed on robots in real and simulation environments.}
\label{fig:teleop}
% https://docs.google.com/drawings/d/1e0JUrXRBu7AnG3yvGNXQK6XGfxZFf9_9RgF2eVlx5TA/edit?usp=sharing
\end{figure}

We use the following real-world robot datasets:
\begin{enumerate}
    \item \textbf{\modelname{} Humanoid Pre-Training Dataset.} Our internally collected dataset covers a broad range of general manipulation tasks, focused on Fourier GR1 through teleoperation. We leverage the VIVE Ultimate Tracker to capture the teleoperator’s wrist poses while Xsens Metagloves track finger movements. We also explored other teleoperation hardware options, including Apple Vision Pro and Leap Motion (see \FIGREF{fig:teleop}).   
    The recorded human movements are then retargeted to humanoid actions via inverse kinematics. The real-time teleoperation operates at a control frequency of 20Hz. Alongside the robot's actions, we capture images from a head-mounted camera at each step, as well as the human’s low-dimensional proprioception and actions. The dataset includes fine-grained annotations, which detail atomic actions such as grasping, moving, and placing, and coarse-grained annotations, which aggregate sequences of fine-grained actions into higher-level task representations. This hierarchical structure supports learning both precise motion control and high-level task reasoning.
    
    \item \textbf{Open X-Embodiment.} \citet{open_x_embodiment_rt_x_2023} is a widely used cross-embodiment dataset for robot manipulation. 
    We include the RT-1~\citep{rt1-2022}, Bridge-v2~\citep{walke2023bridgedata}, Language Table~\citep{lynch2022interactivelanguagetalkingrobots}, DROID~\citep{khazatsky2024droid}, MUTEX~\citep{shah2023mutex}, RoboSet~\citep{roboset} and Plex~\citep{thomas2023plex}, providing diverse datasets covering various manipulation tasks, language-conditioned control, and robot-environment interactions.
    
    \item \textbf{AgiBot-Alpha.} \citet{contributors2025agibotworld} is a large-scale dataset of trajectories from 100 robots. We used the 140,000 trajectories available at the time of launching our training run. The dataset covers fine-grained manipulation, tool usage, and multi-robot collaboration.
\end{enumerate}

\subsection{Synthetic Datasets}
\label{datasets:synthetic-datasets}
Our synthetic datasets include 1) simulation trajectories automatically multiplied from a small number of human demonstrations within physics simulators and 2) neural trajectories derived from videos produced by off-the-shelf neural generation models. 

\paragraph{Simulation Trajectories}
%\checkthis{@Yu and @Soroush, describe 54DC digital cousins and how we collect human demonstrations and use DexMG to generate the trajectories}
In addition to real-world datasets, we feature large-scale synthetic datasets generated in simulation as described in Sec.~\ref{sec:training_data_generation}.
Our simulation tasks comprise humanoid robots performing a broad range of tabletop rearrangement tasks and feature a large array of realistic 3D assets.
We build these tasks under the RoboCasa simulation framework~\citep{robocasa2024}.
Broadly, our tasks follow the behavior ``rearrange A from B to C'', where A corresponds to an object, and B and C represent the source and target locations in the environment.
% We feature dozens of object categories spanning everyday food items.
The source and target locations are receptacles such as plates, baskets, placemats, and shelves, and the robot must rearrange objects across different combinations of source and target receptacles.
Overall, our pre-training simulation datasets feature 54 unique combinations of source and target receptacle categories.
We place the objects and receptacles in randomized locations throughout the table and additionally incorporate distractor objects and receptacles in the scene.
The distractors require the model to pay attention to the task language to perform the desired behavior.

We generate diverse, high-quality training datasets at a massive scale using DexMimicGen.
Our datasets feature the GR-1 humanoid robot, but we can adopt the system for a wide range of robots.
We begin by collecting a few dozen source demonstrations via teleoperation using the Leap Motion device.
The Leap Motion device tracks the 6-DoF wrist poses and finger poses, and we retarget these values and send them to the whole-body IK controller based on mink~\citep{Zakka_Mink_Python_inverse_2024}.
Given human demonstrations, DexMimicGen processes the demonstrations into object-centric segments and then transforms and combines these segments to generate new demonstrations.
Using this system, we generate 10,000 new demonstrations for each (source, target) receptacle pair in our pre-training task regime, resulting in 540k total demonstrations.

\paragraph{Neural Trajectories}
To generate neural trajectories, we fine-tune open-source image-to-video models on our real-world \modelname{} Humanoid Pre-Training dataset, as described in Sec.~\ref{sec:training_data_generation}. 
We trained the models for 100 epochs on a dataset comprising 3,000 real-world robot data samples with language annotations, each recorded at 480P resolution and consisting of 81 frames.
As illustrated in \FIGREF{fig:dream_sample}, our model can generate high-quality counterfactual trajectories given novel language prompts. 
Moreover, the model, trained on Internet-scale video data, demonstrates strong generalization capabilities in handling unseen initial frames, novel objects, and new motion patterns.
These videos are further labeled with latent actions and IDM-based pseudo-actions for model training.
We generate a total of around 827 hours of videos; it takes 2 minutes to generate a one-second video on an L40 GPU, and required approximately 105k L40 GPU hours ($\sim$1.5 days) on 3,600 L40 GPUs. 

% \textbf{@Zhenjia Xu (PIC)}: We also collect a large amount of data on Fourier GR1 using teleoperation. Specifically, we use the VIVE Ultimate Tracker to capture the teleoperator's wrist pose. Additionally, Xsens Metagloves are worn to track finger poses. The tracked data is then retargeted to humanoid actions via inverse kinematics. The control frequency of this real-time teleoperation is 20Hz. In addition to the robot's actions, images from a head-mounted camera are recorded at each step, along with the human's low-dimensional proprioception and actions.

\subsection{Human Video Datasets}
\label{datasets:human-video-datasets}
We include a diverse set of human video datasets.
These do not include explicit action labels but contain extensive sequences of human-object interactions, capturing affordances, task semantics, and natural motion patterns.
These datasets cover a wide range of real-world human behaviors, including grasping, tool use, cooking, assembly, and other task-oriented activities performed in natural environments, and provide detailed first-person perspectives of hand-object interactions (examples shown in Figure \ref{fig:human_sample}). Our video datasets include the following:

\begin{itemize}
    \item \textbf{Ego4D} is a large-scale egocentric video dataset that includes diverse recordings of everyday  activities~\citep{grauman2022ego4d};
    \item \textbf{Ego-Exo4D} adds complementary exocentric (third-person) views alongside first-person recordings~\citep{grauman2024ego};
    \item \textbf{Assembly-101} focuses on complex assembly tasks by providing detailed videos of step-by-step object assembly~\citep{sener2022assembly101};
    \item \textbf{EPIC-KITCHENS} includes first-person footage of culinary activities~\citep{damen2018scaling};
    \item \textbf{HOI4D} captures human-object interactions with frame-wise annotations for segmentation, hand and object poses, and actions~\citep{liu2022hoi4d};
    \item \textbf{HoloAssist} captures collaborative and assistive tasks within augmented reality environments~\citep{wang2023holoassist};
    \item \textbf{RH20T-Human} includes recordings of fine-grained manipulation tasks with an emphasis on natural hand-object interactions across diverse real-world scenarios~\citep{fang2023rh20t}.
\end{itemize}

\section{Evaluation}
\label{sec:experiments}

We evaluate our \modelname{} models in a diverse set of simulated and real-world benchmarks. Our simulation experiments are conducted on three distinct benchmarks designed to systematically assess the effectiveness of our model across various robot embodiments and manipulation tasks. In our real-world experiments, we investigate the model's capability on a suite of tabletop manipulation tasks with the GR-1 humanoid robot. These experiments aim to demonstrate \modelname{}'s ability to acquire new skills from a limited number of human demonstrations.

\begin{figure}[t]
\centering
\includegraphics[width=\textwidth]{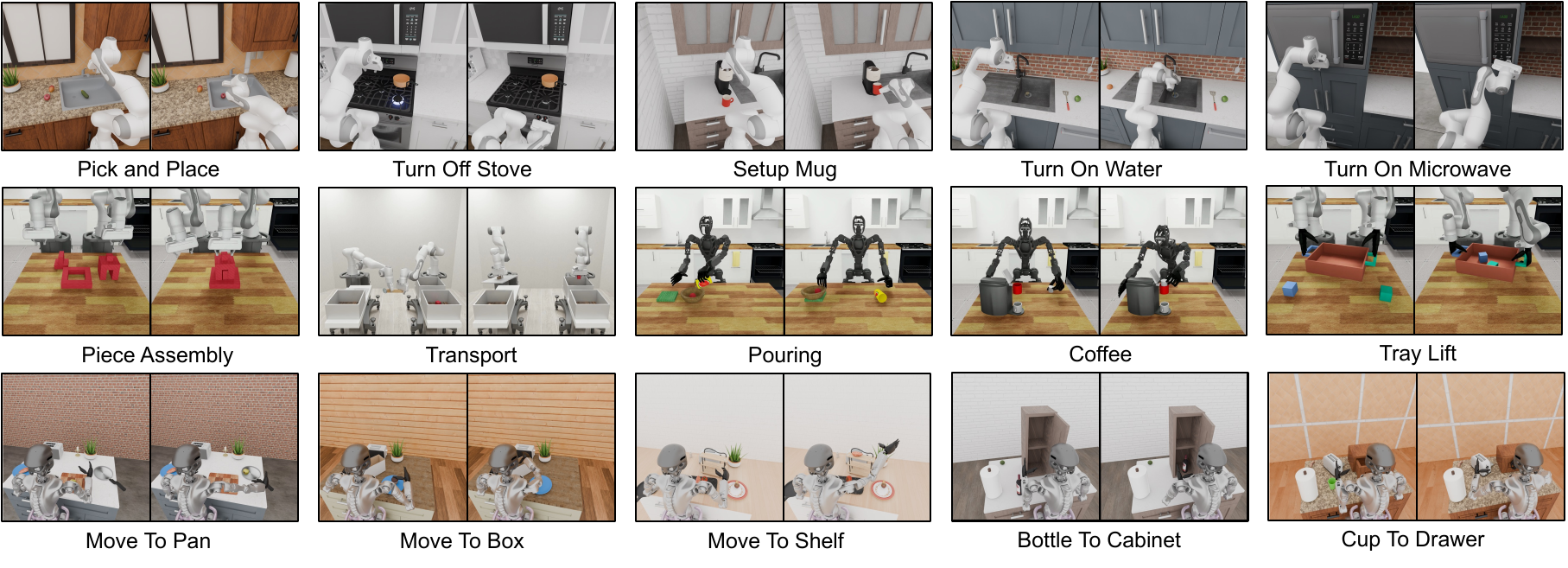}
\caption{\textbf{Simulation Tasks.} Our simulation experiments use tasks from two open-source benchmarks (RoboCasa~\citep{robocasa2024} in the top row and DexMimicGen~\citep{jiang2024dexmimicen} in the middle row) and a newly developed suite of tabletop manipulation tasks that closely resemble our real-world tasks (bottom row). We provide Omniverse renderings of the tasks above.}
\label{fig:sim_tasks}
\end{figure}

\subsection{Simulation Benchmarks}
\label{sec:sim_bench}

Our simulation experiments comprise two open-source benchmarks from prior work~\citep{robocasa2024,jiang2024dexmimicen}, as well as a newly developed suite of tabletop manipulation tasks designed to closely mirror our real-world task settings. We meticulously choose these benchmarks for evaluating our models across different robot embodiments and diverse manipulation tasks. Our model checkpoints, together with the publicly available simulation environments and datasets, ensure the reproducibility of our key results. Fig.~\ref{fig:sim_tasks} illustrates some example tasks from these three benchmarks. 
% \checkthis{@Ajay and @Runyu, please create a double-column figure showcasing example tasks of from the three, similar to Fig 4 in https://arxiv.org/pdf/2410.24185}
% \textbf{@Soroush Nasiriany (PIC)} together with @Yu Fang, @Zhenyu Jiang, @Runyu Ding

\begin{itemize}
    \item \textbf{RoboCasa Kitchen (24 tasks, RoboCasa)}
    
    RoboCasa~\citep{robocasa2024} features a collection of tasks in simulated kitchen environments. We focus on 24 ``atomic'' tasks that involve foundational sensorimotor skills such as pick-and-place, door opening and closing, pressing buttons, turning faucets, and more. For each task, we use the publicly available dataset of 3000 demonstrations featuring the Franka Emika Panda arm, all generated with MimicGen~\citep{mandlekar2023mimicgen}.
    The observation space includes three RGB images captured from cameras positioned on the left, right, and at the wrist. The state representation comprises the position and rotation of both the end-effector and the robot base, as well as the gripper’s state. The action space is defined by the relative position and rotation of the end-effector along with the gripper state.
    We follow the same training and evaluation protocol outlined by~\cite{robocasa2024}.
    
    \item \textbf{DexMimicGen Cross-Embodiment Suite (9 tasks, DexMG)}
    
    DexMimicGen~\citep{jiang2024dexmimicen} includes an array of nine bimanual dexterous manipulation tasks requiring precise two-arm coordination. Together, these tasks cover three bi-manual robot embodiments: (1) \textit{Bimanual Panda Arms with Parallel-Jaw Grippers}: tasks include threading, piece assembly, and transport. The state/action space consists of the end-effector position and rotation of both arms, as well as the gripper states; (2) \textit{Bimanual Panda Arms with Dexterous Hands}: tasks include box cleanup, drawer cleanup, and tray lifting. The state/action space consists of the end-effector position and rotation of both arms and hands; (3) \textit{GR-1 Humanoid with Dexterous Hands}: tasks include pouring, coffee preparation, and can sorting. The state/action space consists of the joint position and rotation of both arms and hands, along with the waist and neck.
    We generate 1000 demonstrations for each task using the DexMimicGen data generation system and evaluate the model's ability to generalize to novel object configurations.
    
    \item \textbf{GR-1 Tabletop Tasks (24 tasks, GR-1)}
    
    This dataset serves as a digital counterpart to real-world humanoid datasets, enabling systematic evaluations that inform the performance of real-robot deployment. This benchmark focuses on dexterous hand control using the GR-1 humanoid robot equipped with Fourier dexterous hands. Compared to DexMG, this benchmark features a significantly larger variety of objects with diverse placements. We model a total of 18 rearrangement tasks, which have a similar structure to the pre-training tasks outlined in 
    Sec.~\ref{datasets:synthetic-datasets}, \ie, rearranging objects from a source to a target receptacle. Each task involves a unique combination of receptacles, and these combinations are unseen in our pre-training data.
    Like the pre-training tasks, most tasks involve distractor objects and receptacles that require the model to pay attention to the task language.
    We additionally feature six tasks that involve placing objects into articulated objects (\ie, cabinets, drawers, and microwaves) and closing them.
    The observation space includes one RGB image captured from an egocentric camera positioned on the robot's head.
    The state/action space consists of the joint position and rotation of both arms and hands, along with the waist and neck. 
    We optionally include in our datasets the end effector-based actions for controlling the arms, as the native action space for controlling the whole-body IK controller is end effector-based.
    We generate 1000 demonstrations for each task using the DexMimicGen system. %data generation system.
    % Camera images and language instruction tokens are the main observations provided to our model for training and inference. Across different embodiments, these factors vary. For example, different robots may have multiple or single views with varying resolutions, and different task categories are expressed in descriptive natural language. In addition to external sensory inputs, proprioception—such as joint positions and velocities—provides crucial information about the robot’s internal state. Based on these observations, the model generates actions to control the robot. Generally, robot controllers operate in two modes: end-effector control and joint position control. Our model supports both modes and can switch between them seamlessly.

\end{itemize}

\subsection{Real-World Benchmarks}
\label{sec:real_bench}
\begin{figure}[ht!]
    \centering
    \includegraphics[width=\linewidth]{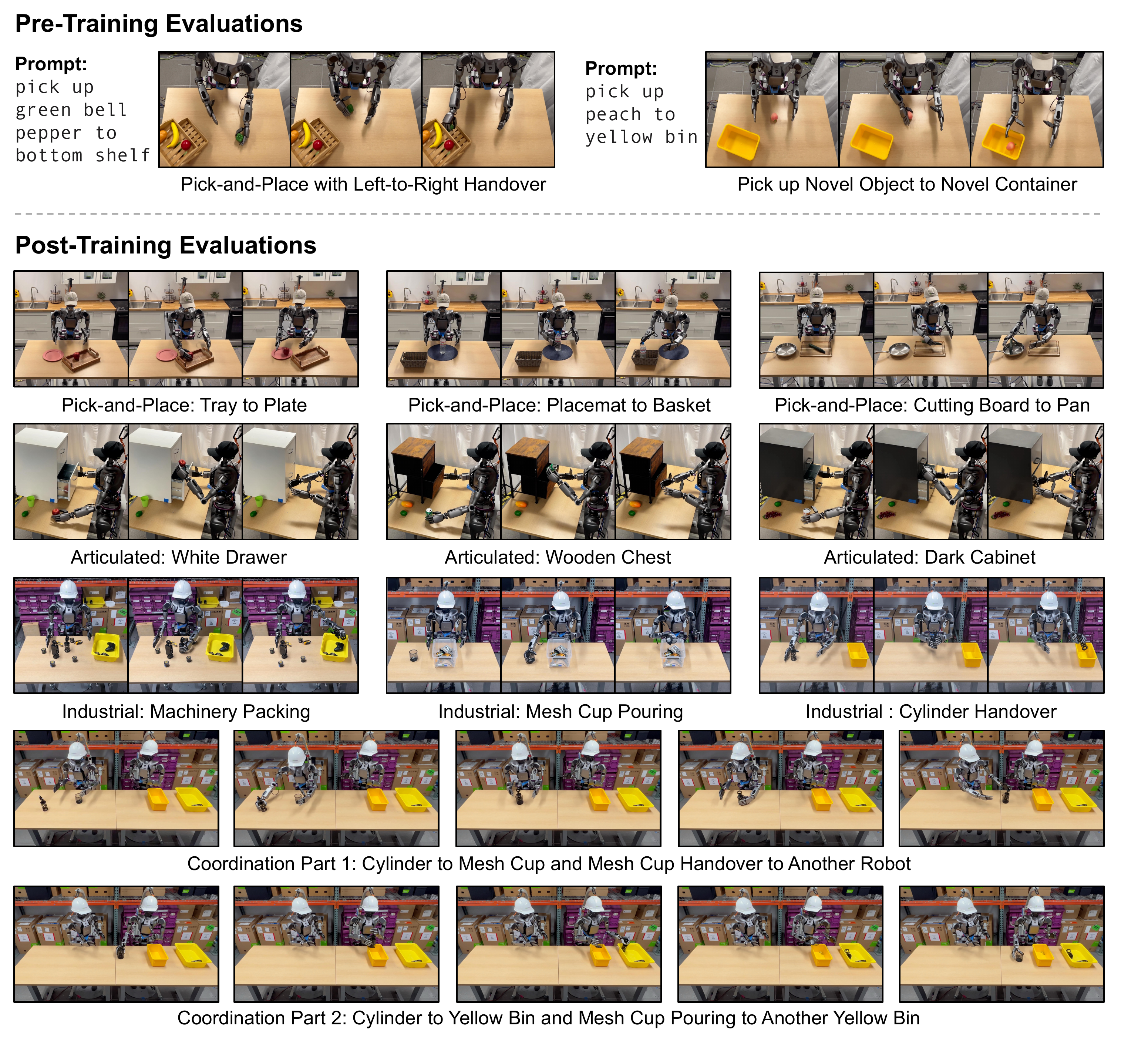}
 \caption{\textbf{Real-World Tasks.} All images are captured from policy rollouts of \modellarge{} and models post-trained from \modellarge{}. \textbf{(Top) Pre-training evaluations.} We design two manipulation tasks to assess our pretrained models. The left image shows a left-to-right handover, while the right image illustrates the placement of novel objects into an unseen target container.
\textbf{(Bottom) Post-training evaluations.} We introduce four distinct task categories. From top to bottom, we present examples of object-to-container pick-and-place, articulated object manipulation, industrial object manipulation, and multi-agent coordination.
    }
\label{fig:real_gr1_task}
\end{figure}

We introduce a diverse and meticulously designed set of tabletop manipulation tasks, aimed at evaluating and post-training our models on human demonstrations. These tasks emphasize critical aspects of real-world dexterity, including precise object manipulation, spatial reasoning, bimanual coordination, and multi-agent collaboration. We carefully categorize our benchmarks into four distinct types, ensuring a rigorous evaluation of model performance. We show some example tasks from our real-world benchmarks in Fig.~\ref{fig:real_gr1_task}.

\begin{itemize}

    \item \textbf{Object-to-Container Pick-and-Place (5 tasks, Pick-and-Place)}
    
    This category evaluates the model’s ability to grasp objects and place them into designated containers, a fundamental capability for robotic manipulation. Tasks include transferring objects between common household containers such as trays, plates, cutting boards, baskets, placemats, bowls, and pans. These scenarios test fine motor skills, spatial alignment, and adaptability to different object geometries. To rigorously assess generalization, we evaluate models on both seen and unseen objects.

    \item \textbf{Articulated Object Manipulation (3 tasks, Articulated)}
    
    These tasks assess the model's ability to manipulate articulated storage compartments. The model must grasp an object, place it into a storage unit such as a wooden chest, dark cabinet, or white drawer, and then close the compartment. These tasks introduce challenges in constrained motion control and precise placement within limited spaces. Generalization is tested with both seen and unseen objects.

    \item \textbf{Industrial Object Manipulation (3 tasks, Industrial)}
    
    We design this category for industrial scenarios, which involve three structured workflows and tool-based interactions: 1) \textit{Machinery Packing}: Pick up various machinery parts and tools and place them into a designated yellow bin; 2) \textit{Mesh Cup Pouring}: Grasp a mesh cup containing small industrial components (\eg, screws and bolts) and pour its contents into a plastic bin; and 3) \textit{Cylinder Handover}: Pick up a cylindrical object, transfer it from one hand to the other, and place it into a yellow bin.
    These tasks closely mirror real-world industrial applications, making them highly relevant benchmarks for assessing dexterity in structured environments.

    \item \textbf{Multi-Agent Coordination (2 tasks, Coordination)}
    
    Collaborative tasks require synchronization between multiple agents, emphasizing role coordination and adaptive decision-making:
    1) \textit{Coordination Part 1}: Pick up a cylinder, place it into a mesh cup, and hand it over to another robot; and 2) \textit{Coordination Part 2}: The receiving robot places the cylinder into one yellow bin, then pours the remaining contents of the mesh cup into another yellow bin.
    
\end{itemize}

These carefully designed benchmarks introduce structured, goal-driven interactions to test whether a model can seamlessly adapt to real-world applications. To build a high-quality post-training dataset, we let human operators collect task-specific data for durations ranging from 15 minutes to 3 hours, depending on task complexity. We then filter out low-quality trajectories to maintain data integrity. By incorporating a diverse set of task requirements --- spanning precise single-agent manipulation to complex multi-agent coordination—our benchmark provides a rigorous testbed for evaluating generalization, adaptability, and fine-tuned control in human-like manipulation tasks.

\subsection{Experiment Setup}
Our evaluation experiment consists of post-training  \modelname{} and baseline models as described in Sec.~\ref{sec:training_details} in a data-limited setting and evaluating the policy success rate in our simulated and real benchmarks described in Sections~\ref{sec:sim_bench} and~\ref{sec:real_bench}, respectively. 
By default we use a global batch size of 1024 and train for 60k steps.
For the DexMimicGen Cross-Embodiment Suite, where each embodiment contains relatively few tasks and the overall training data is limited, we used a smaller batch size of 128 for~\modellarge{}.

\paragraph{\textbf{Baselines}}
To demonstrate the effectiveness of diverse pretraining of \modelname{}, we compare with two established baselines, BC-Transformer~\citep{robomimic2021} and Diffusion Policy~\citep{chi2024diffusionpolicy}. We describe the details of these two methods below:
\begin{itemize}
    \item \textbf{BC-Transformer} is a Transformer-based behavior cloning policy in RoboMimic~\citep{robomimic2021}. It consists of a Transformer architecture for processing observation sequences and a Gaussian Mixture Model (GMM) module for modeling action distributions. The policy takes 10 observation frames as input and predicts the next 10 actions.
    \item \textbf{Diffusion Policy}~\citep{chi2024diffusionpolicy} models action distributions through a diffusion-based generative process. It employs a U-Net architecture that progressively removes noise from random samples to generate precise robot actions conditioned on observation sequences. It takes a single frame of observations as input and produces 16 action steps in one inference pass.
\end{itemize}

\paragraph{Evaluation Protocol} 
For simulated benchmark evaluation, we report the average success rate over 100 trials, taking the maximum score of the last 5 checkpoints, where checkpoints are written every 500 training steps, following the protocol from RoboCasa~\citep{robocasa2024}.

For real robot evaluation, we employ a partial scoring system to capture model behavior across different execution phases, ensuring a fine-grained assessment of performance. We report the average success rate over 10 trials for each task, except for the task of \textit{Pack Machinery}; for this task, we report the success rate of how many objects out of the 5 machinery parts and tools are placed into the bin, given a time-limit of 30 seconds. We conduct only 5 trials due to the time constraint. Additionally, to assess the model’s efficiency in a low-data regime, we subsample 10\% of the full dataset for each task and evaluate whether the model can still learn effective behaviors.

\subsection{Quantitative Results}
\label{sec:quant_results}

\paragraph{Pre-training Evaluations} 
To evaluate the generalization capabilities of our pretrained checkpoint, we design two tasks on the real GR-1 humanoid robot (Fig.~\ref{fig:real_gr1_task}). In the first task, the robot is instructed to place an object on the bottom shelf. However, the object is intentionally positioned to the left of its left hand, requiring a coordinated bimanual strategy. The robot must first grasp the object with its left hand, transfer it within reach of the right hand, and then complete the placement onto the shelf.
In the second task, the robot is instructed to place a novel object into an unseen target container. For each task, we evaluate the pretrained \modellarge{} model using five different objects, with three trials per object. \modellarge{} achieves a success rate of 76.6\% (11.5/15) in the first coordinated setting and 73.3\% (11/15) in the second setting involving novel object manipulation. 0.5 stands for grasping the object correctly but failing to place the object into the container. The high performance under these two evaluation settings illustrates the effectiveness of large-scale pre-training.

% \begin{figure}[!t]
%     \centering
%     \includegraphics[width=\linewidth]{assets/zero_shot.pdf}
%     \caption{Illustration of the two manipulation tasks used to evaluate our pretrained checkpoint. The left image depicts a left-to-right handover, while the right image demonstrates the placement of novel objects into an unseen target container. 
%     Our model achieves a success rate of \textbf{11.5/15} (\textbf{76.6\%}) trials in the first setting and \textbf{11/15} (\textbf{73.3\%}) trials in the second setting. \checkthis{@Guanzhi to update this figure} (0.5 stands for grasping the object correctly but failing to place the object into the container.)} %\checkthis{it is not clear what half a success is. I.e. how can we get 11.5/15?} }
% \end{figure}

\paragraph{Post-training Evaluations} 
In simulation, we compare the quantitative results for our post-trained \modelname{} models against from-scratch baselines in the three simulation benchmarks (Table~\ref{tab:main_quant_sim}). 
For each benchmark, we post-train using 30, 100, and 300 demonstrations per task (24 tasks for RoboCasa, 9 tasks for DexMG, and 24 tasks for GR-1).
We observe that \modelname{} consistently outperforms the baseline models across benchmark tasks and dataset sizes.
In Appendix~\ref{sec:appendix_posttrain_details}, we include the full results and a bar plot (Fig.~\ref{fig:sim_posttraining}) for comparison.

\begin{table}[h!]
\centering
\caption{\textbf{Simulation Results.} Average success rate across three simulation benchmarks, using 100 demonstrations per task. \modelname{} outperforms both baselines, especially on the GR-1 task where it outperforms by more than 17 \%.}
\label{tab:main_quant_sim}
\begin{tabular}{lcccc}
\toprule
     & RoboCasa  & DexMG     & 
GR-1 & Average  \\ \midrule
BC Transformer &          26.3\%         &  53.9\%  &   16.1\%  & 26.4\% \\
Diffusion Policy   &      25.6\%          &  56.1\%&    32.7\%  &  33.4\%\\ %\midrule
\modellarge{}                       &    \textbf{32.1\%}    & \textbf{66.5\%}&  \textbf{50.0\%}  & \textbf{45.0}\%\\ 
\bottomrule
\end{tabular}
\end{table}

\begin{figure}[!thbp]
    \centering
    \includegraphics[width=\linewidth]{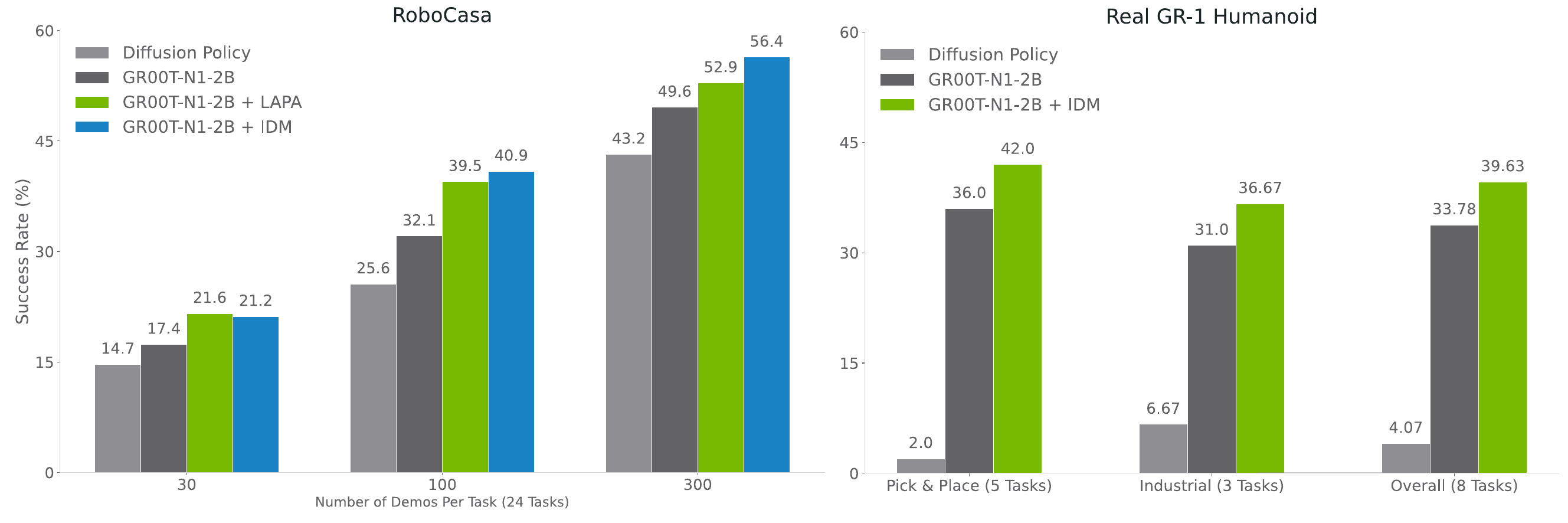}
    \caption{\textbf{Neural Trajectories Ablations}. In the RoboCasa simulation, we show using neural trajectories for post-training across 3 data regimes (30, 100, and 300 per task).  In the real world, we show results only on the low-data regime (10\% of the demonstrations). We co-train with 3k neural trajectories per task for RoboCasa and 100 neural trajectories per task for real-world tasks. We explore using both latent and IDM-labeled actions in simulation and only IDM-labeled actions for the real robot.}
    \label{fig:real_gr1}
\end{figure}

On the real robot, we compare \modellarge{} against Diffusion Policy, training on 10\% of the human teleoperation dataset and the full dataset (Table~\ref{tab:main_quant_real} and \FIGREF{fig:real_gr1}).
\modellarge{}, achieves a significantly higher success rate across all tasks, outperforming Diffusion Policy by 32.4\% in the 10\% Data setting and by 30.4\% in the Full Data setting. Notably, \modellarge{} trained on just 10\% of the data performs only 3.8\% lower than Diffusion Policy trained on the full dataset, highlighting its data efficiency. 

% \begin{figure}[!thbp]
%     \centering
%     \includegraphics[width=\linewidth]{assets/real_gr1.pdf}
%     \caption{Average policy success rate on real-world manipulation tasks with the GR-1 humanoid robots.}
%     \label{fig:real_gr1}
% \end{figure}

\begin{table}[h!]
\centering
\caption{\textbf{Real-World Results.} Average policy success rate on real-world tasks with the GR-1 humanoid robots. \modelname{} beats the diffusion policy baseline and shows strong results even with very little data.}
\label{tab:main_quant_real}
\begin{tabular}{lccccc}
\toprule
                                        & Pick-and-Place      & Articulated & Industrial & Coordination  & Average \\ \midrule
Diffusion Policy (10\% Data)  & 3.0\% & 14.3\% & 6.7\% & 27.5\% & 10.2\% \\
Diffusion Policy (Full Data)  & 36.0\% & 38.6\% & 61.0\%  &  62.5\% & 46.4\%\\ \midrule
\modellarge{} (10\% Data)      & 35.0\% & 62.0\% & 31.0\% & 50.0\% & 42.6\%\\
\modellarge{} (Full Data)      &  \textbf{82.0}\% & \textbf{70.9}\% & \textbf{70.0}\%  & \textbf{82.5}\% & \textbf{76.8}\%\\ 
\bottomrule
\end{tabular}
\end{table}

\paragraph{Post-training w/ Neural Trajectories Evaluations}
We show some preliminary results of using neural trajectories during post-training for the RoboCasa benchmark for simulation evaluation and Pick-and-Place (seen) and Industrial for the real-world evaluation in Figure \ref{fig:real_gr1}. We observe that GR00T N1 co-trained with neural trajectories consistently results in substantial gains compared to GR00T N1 only trained on real-world trajectories: +4.2\%, +8.8\%, +6.8\% on average for 30, 100, and 300 data-regimes, respectively, for RoboCasa and +5.8\% on average across the 8 tasks with the GR-1 Humanoid. 

When comparing LAPA and IDM labels in RoboCasa, an interesting pattern emerges: LAPA slightly outperforms IDM in the relatively low-data regime (30), but as more data becomes available (100 and 300), the performance gap between LAPA and IDM widens. This trend is intuitive— with more data for IDM training, the pseudo-action labels become increasingly aligned with real-world actions, leading to stronger positive transfer. Since GR-1 Humanoid is a relatively ``high-data'' regime for us, we only utilize IDM actions for neural trajectory co-training in the real world.

\subsection{Qualitative Results}
How does this behavior look qualitatively? To answer this, we consider the task ``Turn Sink Spout'' in RoboCasa --- in the 100 sample regime, the DP baseline gets 11.8\% success rate whereas \modelname{} gets 42.2\%. The DP baseline often gets confused about the semantics of the tasks. 
From Table~\ref{tab:main_quant_sim}, we see that \modelname{} has strong results in the low-data regime. It is natural, in the limit of large fine-tuning datasets, that the effect of pre-training dwindles.
% \paragraph{Simulation}
% We illustrate typical trajectories for \modelname{} and the DP baseline in \Cref{fig:qual_sim}. The task is to turn the sink spout to the right. While \modelname{} completes the task, the DP baseline instead attempts to move to the left.

% \checkthis{PIC: @Johan and @Guanzhi}
% \paragraph{Real GR-1 Humanoid}

When prompting the pre-trained \modelname{} model with the task instruction ``Pick up the red apple and place it in the basket,'' one of the tasks in our post-training benchmark, we observe interesting behavioral patterns. In this scenario, we intentionally position the apple to the left of the humanoid hand. Despite seeing few similar tasks during pretraining and exhibiting jerkier motions, the pretrained checkpoint uses its left hand to grasp the apple, hands it over to the right hand, and then places it into the basket. We provide the visualization of this behavior in \FIGREF{fig:gr1-pretraining-qualitative-examples}. In contrast, the post-trained checkpoint fails in this scenario. Since all post-training data exclusively involve the right hand without any inter-hand transfer, the post-trained policy loses the capability to perform this behavior.

For post-trained \modelname{}, we observed that, compared to the baseline Diffusion Policy, its motion is generally much smoother, and its grasping accuracy is significantly higher. 
In contrast, the Diffusion Policy baseline suffers from immobility during the initial frames and frequently exhibits inaccurate grasping, resulting in a low success rate in our real-world benchmarks. We provide visualizations of two policy rollout examples in \FIGREF{fig:gr1-posttraining-qualitative-examples}.

\subsection{Limitations}
Currently, our \modelname{} model focuses primarily on short-horizon tabletop manipulation tasks. In future work, we aim to extend its capabilities to tackle long-horizon loco-manipulation, which will require advancements in humanoid hardware, model architecture, and training corpora. We anticipate a stronger vision-language backbone will enhance the model’s spatial reasoning, language understanding, and adaptability. Our synthetic data generation techniques --- leveraging video generation models and automated trajectory synthesis systems --- have shown great promise. However, existing methods still face challenges in generating diverse and counterfactual data, while adhering to the laws of physics, limiting the quality and variability of synthetic datasets. We aim to enhance our synthetic data generation techniques to further enrich our data pyramid for model training. Furthermore, we plan to explore novel model architectures and pre-training strategies to improve the robustness and generalization capabilities of our generalist robot models.

\section{Related Work}

\textbf{Foundation Models in Robotics.} 
Developing and using foundation models~\citep{bommasani2021opportunities} for robotics has been of great interest recently. 
% non-VLA work that uses LLMs/VLMs 
One common approach is to leverage existing pre-trained foundation models as high-level black-box reasoning modules in conjunction with low-level robot-specific policies~\citep{saycan-2022, innermono-2022, progprompt-2022, driess2023palm, code-as-policies-2022, text2motion-23, groundeddecoding-2024}. This approach allows the robot to plan sequences of low-level skills or motions using the pre-trained foundation model. However, it assumes the availability of these low-level policies and a sufficient interface to connect them to the black-box foundation models. 
% one core limitation of this approach is a lack of end-to-end optimization for the downstream robotics task. 
% VLA work
An alternative approach is to finetune pre-trained foundation models on robotics data to build Vision-Language-Action (VLA) models~\citep{rt1-2022, rt22023arxiv, black2024pi_0, kim24openvla, zheng2025tracevla, wen2024tinyvla, cheang2024gr2vla, li2023vision, zhen20243dvla, huang2023embodied, ye2025latent, yang2025magma}. Instead of enforcing a rigid hierarchy between high-level VLM planning and low-level control, these VLA models allow for end-to-end optimization toward the downstream deployment tasks. We take a similar approach to train \modelname{} and use the Eagle-2 model~\citep{eagle2} as our base Vision Language Model (VLM). We fine-tune our VLM together with a flow-matching~\citep{flowmatching, liu2022flow, hu2024adaflow} action generation model with action chunking~\citep{action_chunking}. In contrast to prior VLA models \citep{black2024pi_0} that use a mixture-of-experts architecture to bridge the base VLM model with the action generation model, we use a simple cross-attention mechanism. This approach provides flexibility regarding the exact architecture of the VLM model and the action generation model we can use. Furthermore, we use embodiment-specific state and action projector modules, which support different robot embodiments, including latent~\citep{ye2025latent} and IDM-based~\citep{baker2022video} actions. The use of these projectors is similar to those in \citet{octo_2023}, though that work did not fine-tune the VLM models.

\textbf{Datasets for Robot Learning.} 
% Collecting large-scale datasets in robotics is a challenging endeavor, but there have been several prior attempts to address this problem. 
A core challenge in robot learning is the scarcity of large-scale, diverse, and embodied datasets necessary to train generalist robots. 
% TODO: we could consider discussing self-supervised approaches here (e.g. RL)
% Real robot teleoperation, large-scale datasets
One common approach is to use robot teleoperation~\citep{zhang2017deep, mandlekar2018roboturk, mandlekar2019scaling, mandlekar2020human, wu2023gello, action_chunking, aldaco2024aloha, fu2024mobile, iyer2024open, dass2024telemoma}, where a human uses a device such as a smartphone or Virtual Reality (VR) controller, to control a robot to perform tasks of interest. 
The robot sensor streams and robot controls during operation are logged to a dataset, allowing for high-quality task demonstrations to be collected.
Recently, this approach has been scaled by utilizing large teams of human operators and robot fleets over extended periods of time (\eg, months), resulting in large-scale robot manipulation datasets with thousands of hours of demonstrations~\citep{ebert2021bridge, brohan2022rt, ahn2022can, lynch2022interactive, o2024open, contributors2025agibotworld, black2024pi_0}. 
However, collecting data this way requires extensive cost and human effort.
Another line of work, instrumented human demonstrations, uses special hardware to capture robot-relevant observation and action data without explicitly teleoperating the target robot. For example, \cite{chi2024universal, seo2024legato} use hand-held robot grippers, \citet{fang2024airexo} uses a robot-like exoskeleton, and \cite{kareer2024egomimicscalingimitationlearning} uses special glasses to capture human hand motions, which are retargeted to robot action data. These approaches tend to result in faster data collection, though they have a mismatch with the downstream robot compared to direct robot teleoperation.
% % Simulation, data generation
% There are other compelling alternatives to the burden of real-world on-robot data collection.
% Recently, several works~\cite{mandlekar2023mimicgen, james2020rlbench, dalal2023imitating, gu2023maniskill2, ha2023scaling, robocasa2024, jiang2024dexmimicen, wang2023robogen, garrett2024skillmimicgen, yang2025physics} have proposed automated data generation pipelines that can leverage simulation to produce thousands of task demonstrations with minimal human effort. 
% This makes it easy to generate large-scale datasets; however, utilizing these datasets can be challenging due to the simulation-to-reality gap.
% % TODO: If we want to cite robot-free data collection like UMI, this could be one place to do it.
% % Dream data
% Another promising avenue is to leverage advances in generative models, such as video generation models, to augment existing sets of robot demonstrations~\cite{agarwal2025cosmos, mandi2022cacti, yu2023scaling, chen2023genaug}. For example, given a language prompt and an initial camera frame that shows a robot workspace, a video generation model could create new video demonstrations for the robot to learn from. Similarly, such a model could be applied to augment the visuals in existing task demonstrations.
% Human video
A separate line of work makes use of human video datasets~\citep{grauman2024ego, grauman2022ego4d, goyal2017something, damen2018scaling, miech2019howto100m}, which are plentiful and substantially easier to collect than on-robot data, as a source of training data for robots. 
Some works~\citep{nair2022r3m, wu2023unleashing, karamcheti2023language} use human video datasets to pre-train representations that are then used as a feature space for training policies on downstream robot datasets. Other works~\cite{bharadhwaj2024gen2act, bharadhwaj2024track2act, ren2025motion} try to jointly use human video data and robot data through intermediate representations for the motions in the video. \citet{ye2025latent} shows that pretraining VLAs with \textit{latent} actions only on human videos yields positive transfer to downstream robotic tasks.
% Prior works~\citep{nair2022r3m, wu2023unleashing, karamcheti2023language, bharadhwaj2024gen2act, bharadhwaj2024track2act, ren2025motion} have attempted to use human videos~\citep{grauman2024ego, grauman2022ego4d, goyal2017something, damen2018scaling, miech2019howto100m} as a source of training data for robots~\checkthis{make this sentence crispier - \textit{how} they use human videos}. 
% While this is a rich source of data, making use of the data can be difficult due to the embodiment gap between humans and robots. In this work, we use embodiment-specific state and action projector models to enable training on these different embodiments.
% \checkthis{also mention instrumented human data like UMI, LEGATO, EgoMimic? KL: added above}
% Model tries to combine all of them together
Rather than relying on a single type of training data, we developed techniques to effectively learn from a diverse assortment of real-world robot data, human video data, and synthetic data.

\textbf{Synthetic Data Generation in Robotics.} 
% Simulation, data generation
Real-world robot data collection requires large amounts of time and considerable human cost. By contrast, data collection in simulation can be substantially more efficient and less painful, making it a compelling alternative. 
% There are other compelling alternatives to the burden of real-world on-robot data collection.
Recently, several works~\citep{mandlekar2023mimicgen, james2020rlbench, dalal2023imitating, gu2023maniskill2, ha2023scaling, robocasa2024, jiang2024dexmimicen, wang2023robogen, garrett2024skillmimicgen, yang2025physics} have proposed automated data generation pipelines that can leverage simulation to produce thousands of task demonstrations with minimal human effort. 
This makes it easy to generate large-scale datasets; however, utilizing these datasets can be challenging due to the simulation-to-reality gap.
% TODO: If we want to cite robot-free data collection like UMI, this could be one place to do it.
% Dream data

Another promising avenue has been using neural generative models to augment existing sets of robot demonstrations~\citep{mandi2022cacti, yu2023scaling, chen2023genaug}. However, previous work have been limited to utilizing in-painting or text-to-image diffusion models to augment the training data. In our work, we leverage the recent advancements in video generative models~\citep{agarwal2025cosmos, wan2.1} to create entire neural trajectories, at a scale that has never been explored before: $\sim$300k neural trajectories which amounts to 827 hours of robot trajectories.

In our model, we make use of large synthetic simulation datasets generated by MimicGen~\citep{mandlekar2023mimicgen} and DexMimicGen~\citep{jiang2024dexmimicen}, as well as neural-generated video datasets with state-of-the-art video generation models. Our way of co-training with synthetically generated and real-world data sets us from other large-scale VLA efforts. 
% In our model, we make use of both synthetically generated simulation data, and data from generative models.~\checkthis{be more concrete how we generate synthetic data? how we do things differently?}

\section{Conclusions}
We have presented \modelname{}, an open foundation model for generalist humanoid robots. \modelname{} features a dual-system model design, leverages heterogeneous training data, and supports multiple robot embodiments. We systematically evaluate it as a generalist policy across simulation benchmarks and on the real GR-1 humanoid robot. Our experiments demonstrate its strong generalization capabilities, enabling robots to learn diverse manipulation skills with high data efficiency. We hope that our open \modellarge{} model, alongside its training datasets and simulation environments, will accelerate the community's progress toward building and deploying generally capable humanoid robots in the wild.

\clearpage
\appendix
\section{Contributors and Acknowledgments}
\label{sec:contributors}

\subsection{Core Contributors}

\paragraph{Model Training}
Scott Reed, Ruijie Zheng, Guanzhi Wang, Johan Bjorck, Joel Jang, Ao Zhang, Jing Wang, Yinzhen Xu, Fengyuan Hu, Seonghyeon Ye, Zongyu Lin, Jiannan Xiang, Loic Magne, Zhiding Yu, Zhiqi Li

\paragraph{Real-Robot and Teleoperation Infrastructure}
Zhenjia Xu, Zu Wang, Xinye (Dennis) Da, Fernando Castañeda, Yizhou Zhao

\paragraph{Real-Robot Experiments}
Guanzhi Wang, Yinzhen Xu, Joel Jang, You Liang Tan, Ruijie Zheng

\paragraph{Simulation Infrastructure}
Yu Fang, Nikita Cherniadev, Runyu Ding, Soroush Nasiriany, Zhenyu Jiang, Kevin Lin, Yuqi Xie

\paragraph{Simulation Experiments}
Soroush Nasiriany, Zhenyu Jiang, Yuqi Xie, Kevin Lin, Yu Fang, Runyu Ding, Nikita Cherniadev, Johan Bjorck, Jing Wang

\paragraph{Video Generation Models and Latent Actions}
Joel Jang, Seonghyeon Ye, Zongyu Lin, Jiannan Xiang

\paragraph{Data Infrastructure and Curation}
Fengyuan Hu, Yinzhen Xu, Avnish Narayan, Loic Magne

\paragraph{Compute Infrastructure and Open-Sourcing}
Avnish Narayan, You Liang Tan, Kaushil Kundalia, Fengyuan Hu

\paragraph{Program Management and Operations}
Qi Wang, Lawrence Lao

\paragraph{Product Lead}
Spencer Huang

\paragraph{Research Leads}
Linxi ``Jim'' Fan, Yuke Zhu

\subsection{Contributors}

Ajay Mandlekar, Jan Kautz, Dieter Fox, Edith Llontop, Hao Zhang, Guilin Liu

\subsection{Acknowledgments}

We thank the 1X team, including Bernt Børnich, Eric Jang, Jorge Milburn, Darien Sleeper, Ralf Mayet, Mohi Khansari, Austin Wang, Vlad Lialin, George Joseph, and Turing Zelsnack for providing support with their humanoid robot hardware and technical support. We thank the Fourier team, including Jie Gu, Roger Cai, Yuxiang Gao, Victor Suen, Hengxin Chen, and Fangzhou Shi, for the hardware support and maintenance of the Fourier GR-1 robots.

We thank Max Fu, Zhengyi Luo, Annika Brundyn, Aastha Jhunjhunwala, Jeff Smith, Yunze Man, Guo Chen, De-An Huang and Xin Dong for their technical discussion and assistance, and Marco Pavone, Soha Pouya, Shiwei Sheng, Di Zeng, Yan Chang, Chirag Majithia, John Welsh, Stephan Pleines, Joydeep Biswas for their feedback on our paper draft.

We thank Erwin Coumans, Billy Okal, John Welsh, Pulkit Goyal, Stephan Pleines, Vishal Kulkarni, Chirag Majithia, Di Zeng, Yan Chang, Soha Pouya, Wei Liu, Rushane Hua, Benjamin Butin, Lionel Gulich, Lakshmi Ramesh, Peter Mcaughan, Piyush Medikeri for internal codebase review and testing.

We thank Kyle Yumen, Jeremy Chimienti, Gianna Calderon, Isabel Zuluaga, Juan Zuluaga, Ivy Tam, Jazmin Sanchez, Jesse Yang, Leilee Naderi, Patrick Lee, Tri Cao, Jenna Diamond, Andrew Mathau, Marina Davila, and Sarah Stoddard for working with us on robot teleoperation data collection and annotation.

We thank Andrew Tao, Padmavathy Subramanian, Bryan Catanzaro and Kari Briski for the support and guidance on commercial VLM training, and Yao Lu, Sean Cha, Zaid Pervaiz Bhat, Subhashree Radhakrishnan, Yilin Zhao, Paris Zhang, Ratnesh Kumar, Vidya Murali, Yao Xu, Osvald Nitski, Dmitry Chichkov, Qing Miao, Elena Lantz, and Jane Polak Scowcroft for their contributions to the commercial VLM data curation.

We thank Arun Shamanna Lakshmi, Eric Colter, Ryan Li, Trasha Dewan, Ethan Yu, Xutong Ren, Fernando Luo, for the OSMO compute infrastructure support, Zhe Zhang for training infrastructure support, Alexis Bjorlin for compute resource support, Amit Goel, Amit, Sandra, Leela Karumbunathan for humanoid robot ecosystem support. We thank Ming-Yu Liu, Yen-Chen Lin, Jinwei Gu, Lyne Tchapmi, Qinsheng Zhang for Cosmos technical support, Madison Huang, Douglas Chang, Kalyan Vadrevu, Oyindamola Omotuyi for marketing support, Sangeeta Subramanian, Shri Sundaram, Vishal Kulkarni for Isaac product support, and Bill Dally and Jensen Huang for their leadership, vision, and guidance.

% \newpage

\section{Detailed Experiment Results}
\label{sec:appendix_posttrain_details}
Table~\ref{tab:performance_comparison} and Table~\ref{tab:real_success_rate_extended} present a detailed per-task comparison of our \modellarge{} and the Diffusion Policy baseline across our simulation benchmarks and real-world benchmarks, respectively. We train both models on datasets of varying sizes --- 30, 100, and 300 demonstrations for simulation benchmarks, and 10\% and full data for real-world benchmarks. As expected, performance improves steadily for both models with increasing dataset sizes.  Meanwhile, our model consistently outperforms the baseline across all benchmarks and dataset sizes, indicating better generalization and sample efficiency.

\begin{figure}[h]
    \centering
    \includegraphics[width=\linewidth]{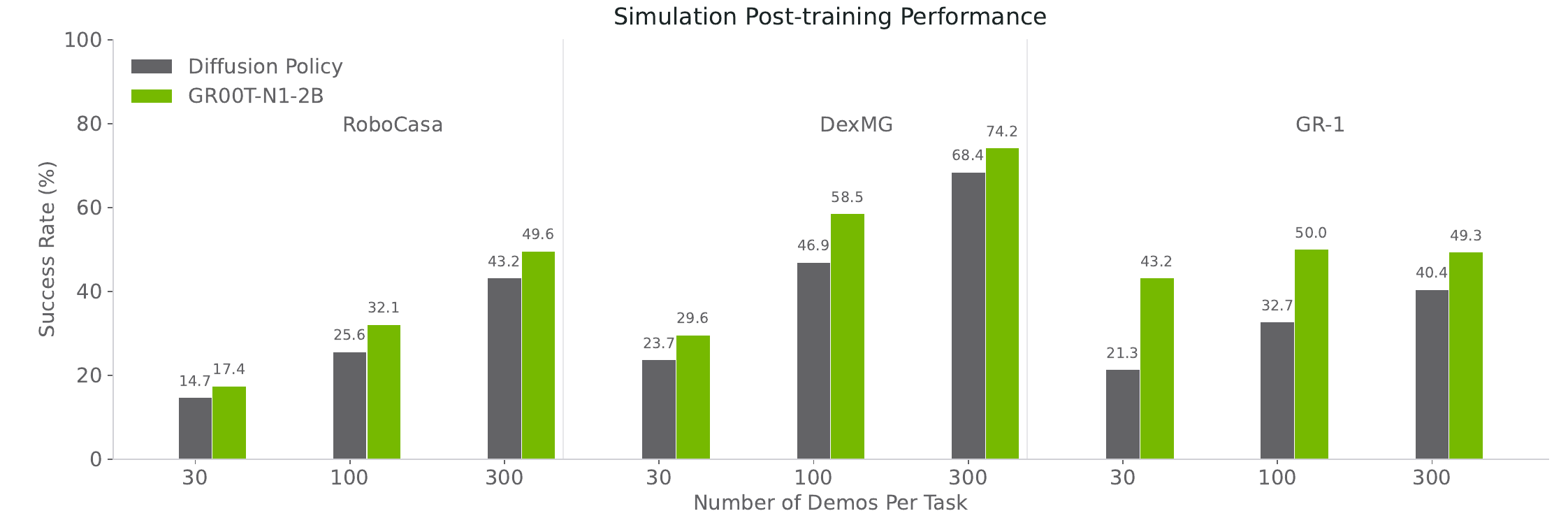}
    \caption{Average policy success rate on simulated manipulation tasks with varying numbers of demonstrations.}
    \label{fig:sim_posttraining}
\end{figure}

\begin{table}[t]
\centering
\footnotesize
\caption{\centering Simulation Evaluation Results with Models Trained with Different Dataset Sizes.}
\label{tab:performance_comparison}
\begin{tabular}{lcccccc}
\toprule
\multirow{2}{*}{Task} & \multicolumn{3}{c}{Diffusion Policy} & \multicolumn{3}{c}{\modellarge{}} \\
\cmidrule(lr){2-4} \cmidrule(lr){5-7}
& 30 demos & 100 demos & 300 demos & 30 demos & 100 demos & 300 demos \\
\midrule
\multicolumn{7}{l}{RoboCasa Kitchen (24 tasks, PnP = Pick-and-Place)} \\
Close Double Door & 1.7 & 26.5 & 60.8 & 0.0 & 43.1 & 74.5 \\
Close Drawer & 57.5 & 88.2 & 94.1 & 76.9 & 96.1 & 99.0 \\
Close Single Door & 21.7 & 46.1 & 72.6 & 49.1 & 67.7 & 83.3 \\
Coffee Press Button & 32.5 & 46.1 & 91.2 & 27.8 & 56.9 & 85.3 \\
Coffee Serve Mug & 6.7 & 28.4 & 66.7 & 3.7 & 34.3 & 72.6 \\
Coffee Setup Mug & 0.0 & 19.6 & 32.4 & 0.0 & 2.0 & 22.6 \\
Open Double Door & 0.0 & 9.8 & 18.6 & 0.0 & 12.8 & 14.7 \\
Open Drawer & 15.8 & 42.2 & 61.8 & 9.3 & 42.2 & 79.4 \\
Open Single Door & 36.7 & 42.2 & 57.8 & 20.4 & 54.9 & 58.8 \\
PnP from Cab to Counter & 2.5 & 4.9 & 9.8 & 0.9 & 3.9 & 19.6 \\
PnP from Counter to Cab & 0.0 & 2.9 & 10.8 & 1.9 & 6.9 & 36.3 \\
PnP from Counter to Microwave & 0.0 & 2.0 & 8.8 & 0.0 & 0.0 & 12.8 \\
PnP from Counter to Sink & 0.0 & 0.0 & 13.7 & 0.0 & 1.0 & 9.8 \\
PnP from Counter to Stove & 0.0 & 1.0 & 17.7 & 0.0 & 0.0 & 23.5 \\
PnP from Microwave to Counter & 0.0 & 2.0 & 11.8 & 0.0 & 0.0 & 15.7 \\
PnP from Sink to Counter & 4.2 & 8.8 & 42.2 & 0.0 & 5.9 & 33.3 \\
PnP from Stove to Counter & 1.7 & 2.9 & 23.5 & 0.0 & 0.0 & 29.4 \\
Turn Off Microwave & 63.3 & 53.9 & 52.0 & 47.2 & 57.8 & 70.6 \\
Turn Off Sink Faucet & 21.7 & 63.7 & 72.6 & 49.1 & 67.7 & 72.6 \\
Turn Off Stove & 5.0 & 10.8 & 19.6 & 4.6 & 15.7 & 26.5 \\
Turn On Microwave & 30.0 & 51.0 & 75.5 & 55.6 & 73.5 & 78.4 \\
Turn On Sink Faucet & 31.7 & 27.5 & 63.7 & 33.3 & 59.8 & 62.8 \\
Turn On Stove & 12.5 & 22.6 & 36.3 & 14.8 & 25.5 & 55.9 \\
Turn Sink Spout & 8.3 & 11.8 & 23.5 & 24.1 & 42.2 & 52.9 \\
\midrule
\textbf{RoboCasa Average} & \textbf{14.7} & \textbf{25.6} & \textbf{43.2} & \textbf{17.4} & \textbf{32.1} & \textbf{49.6} \\
\midrule
\multicolumn{7}{l}{DexMimicGen Cross-Embodiment Suite (9 tasks)} \\
Can Sort & 82.8 & 93.1 & 99.4 & 94.8 & 98.0 & 98.0 \\
Coffee & 35.5 & 68.1 & 79.7 & 44.9 & 79.4 & 73.5 \\
Pouring & 37.0 & 62.3 & 68.8 & 54.4 & 71.6 & 87.3 \\
Threading & 4.2 & 18.3 & 27.5 & 3.9 & 37.3 & 60.8 \\
Three Piece Assembly & 10.0 & 32.5 & 63.3 & 10.8 & 43.1 & 69.6 \\
Transport & 7.5 & 25.0 & 53.3 & 7.8 & 48.0 & 61.8 \\
Box Cleanup & 30.0 & 80.8 & 97.5 & 33.3 & 29.4 & 95.1 \\
Drawer Cleanup & 1.7 & 16.7 & 52.5 & 10.8 & 42.2 & 55.9 \\
Lift Tray & 5.0 & 25.0 & 73.3 & 5.8 & 77.5 & 65.7 \\
\midrule
\textbf{DexMG Average} & \textbf{23.7} & \textbf{46.9} & \textbf{68.4} & \textbf{29.6} & \textbf{58.5} & \textbf{74.2} \\
\midrule
\multicolumn{7}{l}{GR-1 Tabletop (24 Tasks)} \\
Cutting Board to Pot & 22.6 & 37.3 & 48.0 & 58.8 & 57.8 & 57.8 \\
Cutting Board to Basket & 19.6 & 42.2 & 29.4 & 43.1 & 61.8 & 56.9 \\
Cutting Board to Tiered Basket & 13.7 & 13.7 & 18.6 & 13.7 & 23.5 & 34.3 \\
Cutting Board to Pan & 28.4 & 48.0 & 57.8 & 67.7 & 65.7 & 68.6 \\
Cutting Board to Cardboard Box & 11.8 & 15.7 & 22.6 & 31.4 & 30.4 & 33.3 \\
Placemat to Bowl & 14.7 & 18.6 & 23.5 & 31.4 & 39.2 & 39.2 \\
Placemat to Plate & 15.7 & 23.5 & 37.3 & 33.3 & 37.3 & 49.0 \\
Placemat to Basket & 15.7 & 25.5 & 41.2 & 50.0 & 46.1 & 55.9 \\
Placemat to Tiered Shelf & 6.9 & 5.9 & 11.8 & 11.8 & 21.6 & 19.6 \\
Plate to Pan & 13.7 & 17.7 & 35.3 & 35.3 & 48.0 & 52.9 \\
Plate to Cardboard Box & 12.8 & 13.7 & 27.5 & 34.3 & 38.2 & 32.4 \\
Plate to Bowl & 15.7 & 18.6 & 31.4 & 41.2 & 42.2 & 34.3 \\
Plate to Plate & 25.5 & 39.2 & 61.8 & 72.6 & 85.3 & 68.6 \\
Tray to Tiered Shelf & 2.0 & 6.9 & 15.7 & 17.7 & 27.5 & 14.7 \\
Tray to Tiered Basket & 12.8 & 34.3 & 39.2 & 33.3 & 49.0 & 45.1 \\
Tray to Plate & 26.5 & 41.2 & 49.0 & 53.9 & 68.6 & 62.8 \\
Tray to Cardboard Box & 21.6 & 37.3 & 40.2 & 51.0 & 55.9 & 54.9 \\
Tray to Pot & 21.6 & 48.0 & 52.9 & 52.0 & 59.8 & 65.7 \\
Wine to Cabinet & 43.1 & 55.9 & 60.8 & 57.8 & 53.9 & 62.8 \\
Place Bottle to Cabinet & 40.2 & 62.8 & 60.8 & 60.8 & 81.4 & 74.5 \\
Place Milk to Microwave & 37.3 & 41.2 & 51.0 & 42.2 & 58.8 & 49.0 \\
Potato to Microwave & 17.7 & 30.4 & 41.2 & 30.4 & 26.5 & 34.3 \\
Cup to Drawer & 24.5 & 32.4 & 36.3 & 36.3 & 44.1 & 40.2 \\
Can to Drawer & 48.0 & 74.5 & 75.5 & 77.5 & 76.5 & 75.5 \\
\midrule
\textbf{GR-1 Average} & \textbf{21.3} & \textbf{32.7} & \textbf{40.4} & \textbf{43.2} & \textbf{50.0} & \textbf{49.3} \\
\bottomrule
\end{tabular}
\end{table}

\begin{table}[htbp]
\centering
\footnotesize
\caption{\centering Success rate on real-world tasks with the GR-1 humanoid robot.}
\label{tab:real_success_rate_extended}
\begin{tabular}{lcccc}
\toprule
 \multirow{2}{*}{Task} & \multicolumn{2}{c}{Diffusion Policy} & \multicolumn{2}{c}{\modellarge{}} \\
\cmidrule(lr){2-3} \cmidrule(lr){4-5}
 & 10\% Data & Full Data & 10\% Data & Full Data \\
\midrule
Tray to Plate & 0.0\% & 20.0\% & 40.0\% & 100.0\% \\
Cutting Board to Basket & 0.0\% & 30.0\% & 10.0\% & 100.0\% \\
Cutting Board to Pan & 0.0\% & 60.0\% & 60.0\% & 80.0\% \\
Plate to Bowl & 0.0\% & 40.0\% & 30.0\% & 100.0\% \\
Placemat to Basket & 10.0\% & 60.0\% & 40.0\% & 80.0\% \\
\midrule
\textbf{Pick-and-Place Seen Object Average} & 2.0\% & 42.0\% & 36.0\% & 92.0\% \\
\midrule
Tray to Plate & 0.0\% & 20.0\% & 30.0\% & 80.0\% \\
Cutting Board to Basket & 10.0\% & 20.0\% & 60.0\% & 60.0\% \\
Cutting Board to Pan & 0.0\% & 40.0\% & 40.0\% & 80.0\% \\
Plate to Bowl & 0.0\% & 20.0\% & 10.0\% & 40.0\% \\
Placemat to Basket & 10.0\% & 50.0\% & 30.0\% & 100.0\% \\
\midrule
\textbf{Pick-and-Place Unseen Object Average} & 4.0\% & 30.0\% & 34.0\% & 72.0\% \\
\midrule
\textbf{Pick-and-Place Average} & \textbf{3.0\%} & \textbf{36.0\%} & \textbf{35.0\%} & \textbf{82.0\%} \\
\midrule[1.5pt] % Thicker border for highlighted rows
White Drawer & 6.6\% & 36.4\% & 26.4\% & 79.9\% \\
Dark Cabinet & 0.0\% & 46.2\% & 86.6\% & 69.7\% \\
Wooden Chest & 36.4\% & 33.2\% & 72.9\% & 63.2\% \\
\midrule
\textbf{Articulated Average} & \textbf{14.3\%} & \textbf{38.6\%} & \textbf{62.0\%} & \textbf{70.9\%} \\
\midrule[1.5pt]
Machinery Packing & 20.0\% & 44.0\% & 8.0\% & 56.0\% \\
Mesh Cup Pouring & 0.0\% & 62.5\% & 65.0\% & 67.5\% \\
Cylinder Handover & 0.0\% & 76.5\% & 20.0\% & 86.6\% \\
\midrule
\textbf{Industrial Average} & \textbf{6.7\%} & \textbf{61.0\%} & \textbf{31.0\%} & \textbf{70.0\%} \\
\midrule[1.5pt]
Coordination Part 1 & 45.0\% & 65.0\% & 70.0\% & 80.0\% \\
Coordination Part 2 & 10.0\% & 60.0\% & 30.0\% & 85.0\% \\
\midrule
\textbf{Coordination Average} & \textbf{27.5\%} & \textbf{62.5\%} & \textbf{50.0\%} & \textbf{82.5\%} \\
\midrule[1.5pt] % Thicker border for highlighted rows
\textbf{Average} & \textbf{10.2\%} & \textbf{46.4\%} & \textbf{42.6\%} & \textbf{76.8\%} \\
\bottomrule
\end{tabular}
\end{table}

\section{Additional Qualitative Results}
\begin{figure}[t!]
    \centering
    \includegraphics[width=\linewidth]{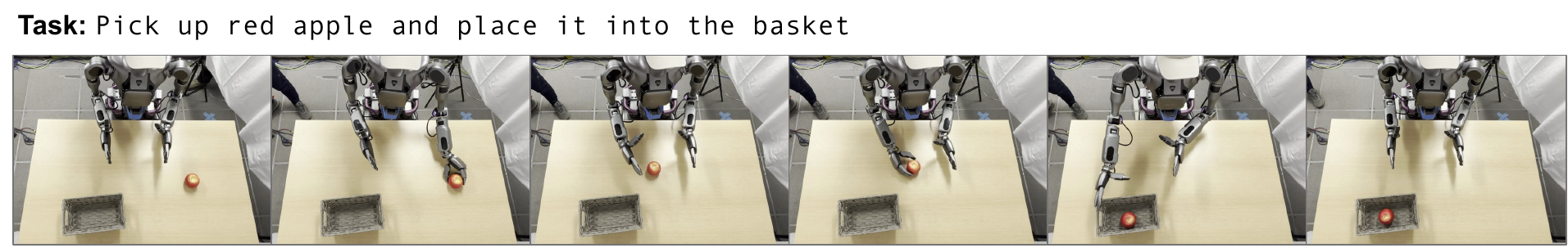}
    \caption{\textbf{Pre-training Qualitative Example.} While prompting the pretrained \modellarge{} model with a post-training task instruction, we even increase the difficulty by placing the apple to the left of both hands. Despite not having encountered this setup during training, the model successfully places the red apple into the basket via a two-handed handover, albeit with jerkier motion.}
    \label{fig:gr1-pretraining-qualitative-examples}
\end{figure}

\begin{figure}[t!]
    \centering
    \includegraphics[width=\linewidth]{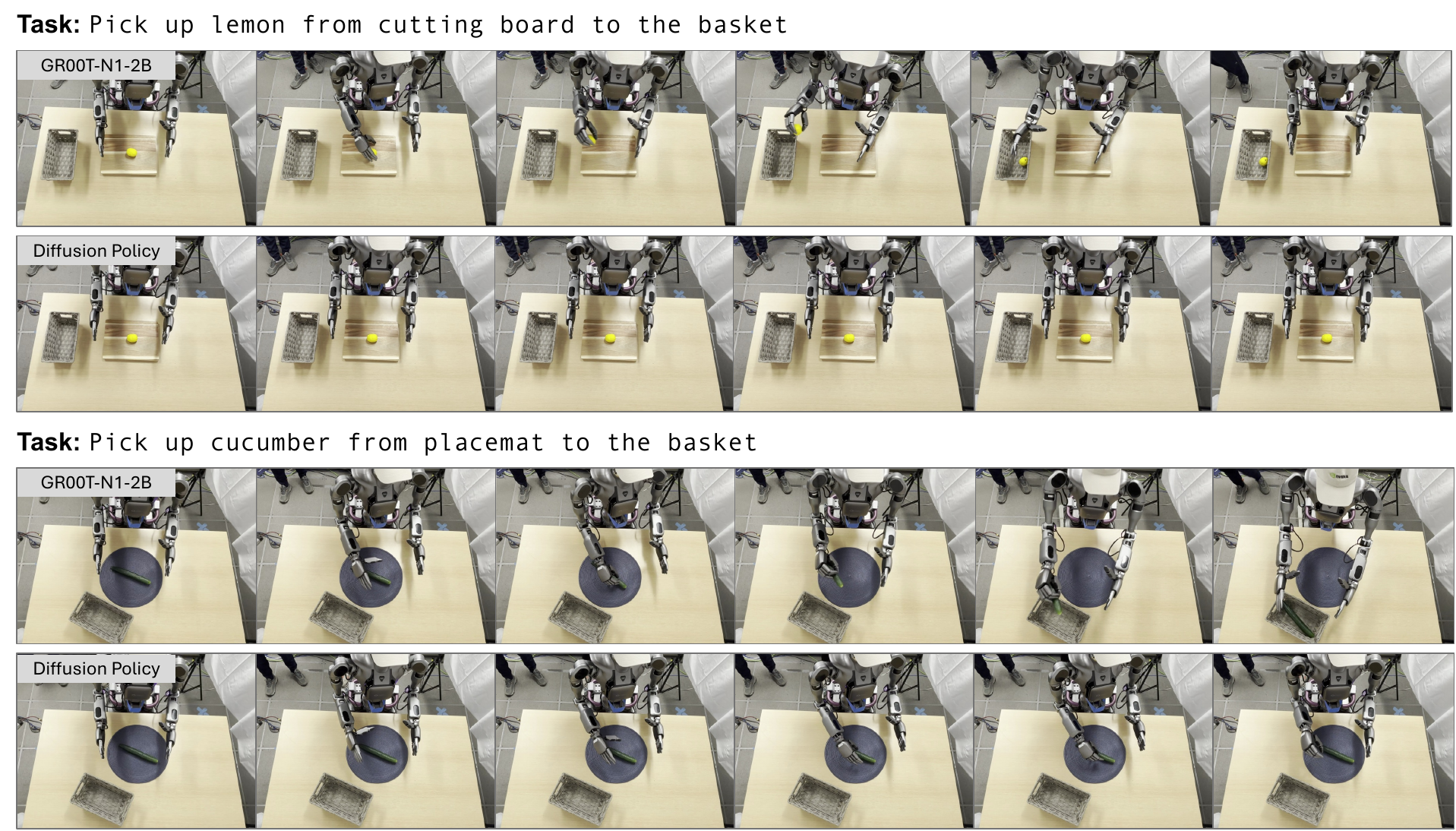}
    \caption{\textbf{Post-training Qualitative Example.} (Top) Post-trained \modellarge{} successfully places the cucumber into the basket, whereas the Diffusion Policy fails due to an inaccurate grasp. (Bottom) The post-trained model successfully picks the lemon from the cutting board and puts it into the pan while the Diffusion Policy remains stuck.}
    \label{fig:gr1-posttraining-qualitative-examples}
\end{figure}

We provide qualitative rollout examples of our pre-trained  and post-trained \modellarge{} model in \FIGREF{fig:gr1-pretraining-qualitative-examples} and \FIGREF{fig:gr1-posttraining-qualitative-examples}. Our model demonstrates strong language following abilities and generalization to unseen situations in bimanual manipulation tasks. %compared to the Diffusion Policy baseline.

\begin{figure}
    \centering
    \includegraphics[width=.9\linewidth]{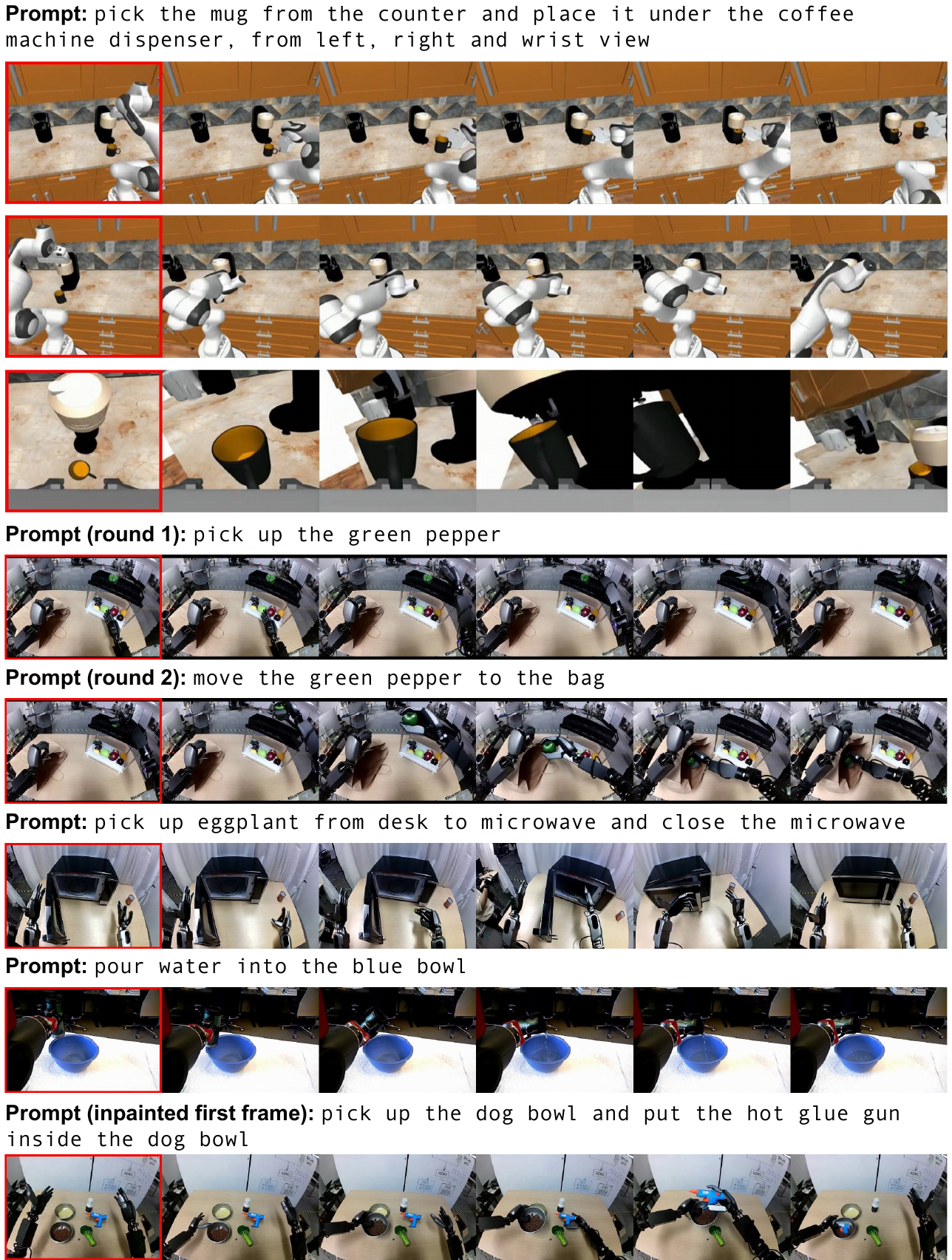}
    \caption{\textbf{More Examples of Neural Generated Trajectories.} We highlight 4 key capabilities of neural trajectories: (1) The first three rows shows an example of a multi-view trajectory generated for the post-training in RoboCasa; we achieve this by concatenating the views as a 4-grid video during training. (2) The following two rows show two consecutive sequences, where the initial frame of the latter is from the end of the former, highlighting the possibility of generating trajectories of tasks requiring composition of atomic tasks. (3)  The following two rows illustrate the capability of our models to generate trajectories with articulated objects and liquids, which are known to be very challenging in simulation. (4) The last row is generated from an in-painted initial frame, showcasing that we can generate more diverse videos without having to collect novel initial frames in the real-world and be bound by human labor. We use the red rectangles to indicate the initial frames.}
    \label{fig:appendix-dream}
\end{figure}

\section{Hyperparameters}
We report important hyperparameters used for the pre- and post-training phases in Table~\ref{tab:hyperparameters}. Overall, these two phases share the same values for most of the hyperparameters. 
For post-training, we use smaller batch sizes to avoid overfitting when fine-tuning in data-limited settings.

\begin{table}[htb]
\centering
\caption{\centering{Training hyperparameters. Pre- and post-training use the same hyperparameters unless specified.}}
\begin{tabular}{lll}
\toprule
\textbf{Hyperparameter} & \textbf{Pre-training Value} & \textbf{Post-training Value}\\
\midrule
Learning rate & 1e-4 \\
Optimizer & AdamW \\
Adam beta1 & 0.95 \\
Adam beta2 & 0.999 \\
Adam epsilon & 1e-8 \\
Weight decay & 1e-5 \\
LR scheduler & cosine \\
Warmup ratio & 0.05 \\
Batch size & 16,384 & 128 or 1024\\
Gradient steps &  200,000 & 20,000 -- 60,000 \\
Backbone's vision encoder & unfrozen \\
Backbone's text tokenizer & frozen \\
DiT & unfrozen \\
\bottomrule
\end{tabular}
\label{tab:hyperparameters}
\end{table}

\section{System Design}

\subsection{Dataset Formats}

Our training corpora build upon the LeRobot dataset format~\citep{cadene2024lerobot}, a widely adopted standard in the open-source robotics community. Developed by Hugging Face, LeRobot aims to lower the barrier to entry for robotics research by providing a standardized format for storing, sharing, and utilizing robot demonstration data. The format has gained significant traction due to its flexibility and the extensive collection of pretrained models and datasets available through the Hugging Face hub.

At its core, the LeRobot dataset format employs a combination of established file formats for efficient storage and access:

\begin{enumerate}
    \item \textbf{Tabular Data:} Robot states, actions, and metadata are stored in parquet files, which offer efficient columnar storage and fast data retrieval. This format enables quick filtering and slicing operations essential for training deep learning models.
    
    \item \textbf{Image and Video Data:} Visual observations are encoded as MP4 video files (or alternatively as PNG image sequences), with references stored in the parquet files. This approach significantly reduces storage requirements while maintaining data accessibility.
    
    \item \textbf{Metadata:} Dataset statistics, episode indices, and other metadata are stored in structured JSON files, providing machine-readable information about the dataset's characteristics.
\end{enumerate}

The format organizes demonstration data into episodes, with each frame containing synchronized observation and action pairs. Each observation typically includes camera imagery (\texttt{observation.images.*}) and robot state information (\texttt{observation.state}), while actions represent the control commands sent to the robot. This organization facilitates both imitation learning, where models learn to predict actions from observations, and reinforcement learning, where models learn to optimize for specific outcomes.

While the LeRobot format provides a solid foundation, our work with cross-embodiment data necessitated additional structure to support richer modality information and more sophisticated training regimes. We have extended the LeRobot format with the following constraints:

\begin{enumerate}
    \item \textbf{Modality configuration file:} We require a \texttt{modality.json} configuration file in the \texttt{meta} directory that explicitly defines the structure of state and action vectors, mapping each dimension to a semantic meaning and provides additional modality-specific information.
    
    \item \textbf{Fine-grained modality specification:} Unlike the standard LeRobot format, which treats state and action as monolithic vectors, our extension splits these vectors into semantically meaningful fields (\eg, end-effector position, orientation, gripper state), each with their own metadata including data types, ranges, and transformation specifications.
    
    \item \textbf{Multiple annotation support:} We extend the format to support multiple annotation types (\eg, task descriptions, validity flags, success indicators) within a single dataset, following the LeRobot convention of storing indices in the parquet file with the actual content in separate JSON files.
    
    \item \textbf{Rotation type specification:} Our format explicitly specifies the representation used for rotational data (\eg, quaternions, Euler angles, axis-angle), enabling proper handling of rotational transformations during training.
\end{enumerate}

Our extended format offers several key benefits for training VLA models:

\begin{enumerate}
    \item \textbf{Semantic clarity:} By explicitly defining the structure and meaning of each dimension in state and action vectors, our format enhances interpretability and reduces errors during data preprocessing and model training.
    
    \item \textbf{Flexible transformations:} The fine-grained modality specification enables sophisticated, field-specific normalization and transformation during training. For example, rotational data can be properly normalized and augmented according to its specific representation.
    
    \item \textbf{Multi-modal learning support:} The extended format naturally accommodates the diverse data types required for VLA models, including visual observations, state information, action commands, and language annotations, while maintaining clear relationships between these modalities.
    
    \item \textbf{Improved data validation:} The explicit structure enables more thorough validation of datasets, reducing the risk of training with malformed or inconsistent data.
    
    \item \textbf{Enhanced interoperability:} While adding constraints, our format maintains backward compatibility with the LeRobot ecosystem, allowing us to leverage existing tools and datasets while enabling more sophisticated modeling approaches.
\end{enumerate}

The extended format strikes a balance between standardization and flexibility, providing a clear structure for common robotics data while accommodating the specific needs of VLA models. This approach has proven valuable in our work, enabling more efficient training and improved model performance while maintaining compatibility with the broader robotics research community.

\subsection{Standardized Action Spaces}

For the above datasets, we make a \textbf{best-effort unification} of action and state spaces to ensure consistency across different embodiments and control modalities. Several key practices are applied to achieve this standardization:

\begin{enumerate}
    \item \textbf{End-effector rotation state normalization:} State end-effector rotations are converted to a \textit{6D rotation representation} to avoid singularities and discontinuities in traditional Euler angles.
    \item \textbf{End-effector rotation action standardization:} End-effector rotation actions are expressed in \textit{axis-angle representation}, providing a compact and smooth parameterization for rotation control.
    \item \textbf{State and action scaling:} \textit{Min-max normalization} is applied to joint states, joint actions, end-effector state positions, and end-effector action positions and rotations, ensuring uniform value ranges across different robots.
    \item \textbf{Consistent ordering:} The arrangement of state and action vectors follows a standardized sequence: \textit{end-effector rotation, end-effector position, and gripper closeness}, ordered from the \textit{left arm to the right arm} (if applicable).
\end{enumerate}

\section{Additional Training Details}
\paragraph{Auxiliary Object Detection Loss}

To enhance the model's spatial understanding, we introduce an auxiliary object detection loss during training. In addition to predicting actions, the model must also localize the object of interest based on the given language instruction. Specifically, for each frame in a trajectory segment, we annotate the bounding box of the target object using the OWL-v2 object detector~\citep{minderer2023owl}. We then compute the normalized center coordinates of the bounding box, $x_{gt}$, by dividing its $x$ and $y$ coordinates by the image width and height, respectively. 
To predict the 2D coordinates, we append a linear layer atop the final vision-language embedding tokens and optimize using a squared loss: $ L_{det} = \| \textbf{x}_{pred} - \textbf{x}_{gt} \|^2$.
Thus, the final loss is given by: $L = L_{fm} + L_{det}$.

\paragraph{Neural Trajectory Generation}
\label{sec:appendix_dream}
We finetune WAN2.1-I2V-14B~\citep{wan2.1} using LoRA~\citep{hu2022lora} on collected teleoperation trajectories. The trajectories are uniformly downsampled to 81 frames at 480P resolution for finetuning. The resulting image-to-video model generates neural trajectories that capture all possible ``counterfactual scenarios'' in the real world. To ensure quality, we filter out generated videos that do not accurately follow the given language instructions. Specifically, we sample 8 frames from each video and prompt a commercial-grade multimodal LLM to assess whether it adheres to the instructions. Videos that fail this criterion undergo re-captioning, with the videos downsampled to 16 frames at 256P resolution for this process.

\paragraph{IDM Model Training}
\label{sec:appendix_idm}
We train an inverse dynamics model (IDM) by conditioning on two images (current and the future frame) within a trajectory and train to generate action chunks between the two image frames. From preliminary experiments, we observed that adding state information or more image frames did not significantly improve the action prediction performance on the validation set. For the IDM model architecture, we use the Diffusion Transformer module (System 1) with SigLIP-2 vision embeddings and train with a flow-matching objective. We train the IDM model for each embodiment for 30K or 60K, depending on the size of the training set. After training, we pseudo-label the actions given the two images (with the same action horizon as training) for each step of the neural trajectories. 

\begin{figure}[thb!]
\centering
\includegraphics[width=0.95\textwidth]{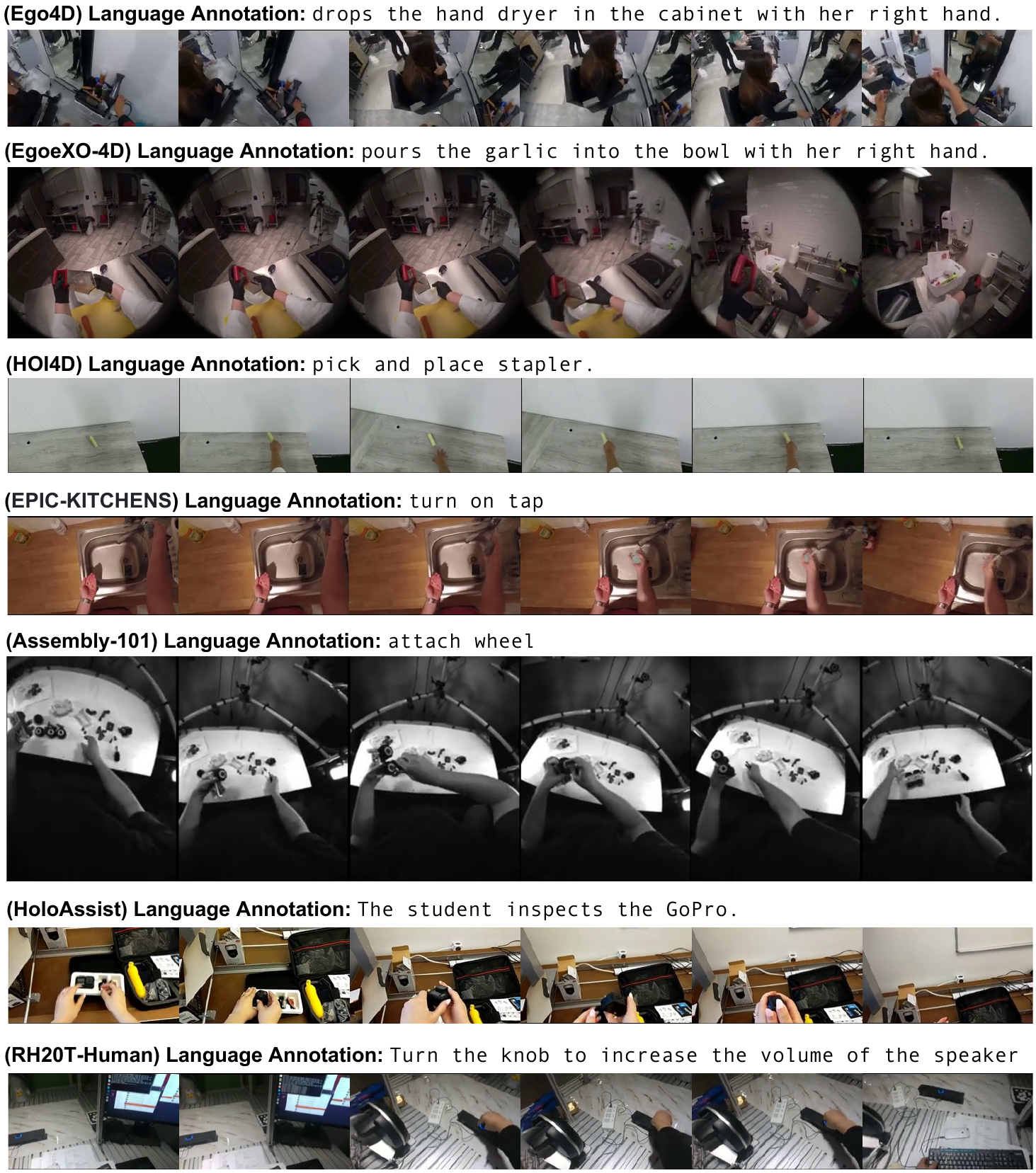}
% \vspace{-8em}
\caption{\textbf{Human Egocentric Video Dataset Samples.} We use seven human video datasets for pre-training. The images above show examples from each of the seven datasets with their corresponding language annotations.}
\label{fig:human_sample}
\end{figure}

\begin{table}[htbp]
\centering
\footnotesize
\caption{\centering Pre-training Dataset Statistics}
\begin{tabular}{llllll}
\toprule
Dataset & Length (Frames) & Duration (hr) & FPS & Camera View & Category \\ 
\midrule
GR-1 Teleop Pre-Training & 6.4M & 88.4 & 20 & Egocentric & Real robot \\
DROID (OXE) & 23.1M & 428.3 & 15 & Left, Right, Wrist & Real robot \\
RT-1 (OXE) & 3.7M & 338.4 & 3 & Egocentric & Real robot \\
Language Table (OXE) & 7.0M & 195.7 & 10 & Front-facing & Real robot \\
Bridge-v2 (OXE) & 2.0M & 111.1 & 5 & Shoulder, left, right, wrist & Real robot \\
MUTEX (OXE) & 362K & 5.0 & 20 & Wrist & Real robot \\
Plex (OXE) & 77K & 1.1 & 20 & Wrist & Real robot \\
RoboSet (OXE) & 1.4M & 78.9 & 5 & Left, Right, Wrist & Real robot \\
Agibot-Alpha & 213.8M & 1,979.4 & 30 & Egocentric, left, right & Real robot \\
RH20T-Robot & 4.5M & 62.5 & 20 & Wrist & Real robot \\
Ego4D & 154.4M & 2,144.7 & 20 & Egocentric & Human \\
Ego-Exo4D & 8.9M & 123.0 & 30 & Egocentric & Human \\
Assembly-101 & 1.4M & 19.3 & 20 & Egocentric & Human \\
HOI4D & 892K & 12.4 & 20 & Egocentric & Human \\
HoloAssist & 12.2M & 169.6 & 20 & Egocentric & Human \\
RH20T-Human & 1.2M & 16.3 & 20 & Egocentric & Human \\
EPIC-KITCHENS & 2.3M & 31.7 & 20 & Egocentric & Human \\
GR-1 Simulation Pre-Training & 125.5M & 1,742.6 & 20 & Egocentric & Simulation  \\
GR-1 Neural Videos & 23.8M & 827.3 & 8 & Egocentric & Neural-generated \\
\midrule
Total robot data & 262.3M & 3,288.8 & -- & -- & -- \\ 
Total human data & 181.3M & 2,517.0 & -- & -- & -- \\
Total simulation data & 125.5M & 1,742.6 & -- & -- & -- \\
Total neural data & 23.8M &  827.3 & -- & -- & -- \\
\bottomrule
Total & 592.9M & 8,375.7 & -- & -- & -- \\ 
\bottomrule
\end{tabular}
\label{tab:dataset_stats}
\end{table}

\clearpage
\setcitestyle{numbers}
\bibliographystyle{plainnat}
\bibliography{main}

\end{document}